\documentclass[twocolumn]{bmcart}

\usepackage{amsthm,amsmath}
\usepackage[implicit=false]{hyperref}
\usepackage[utf8]{inputenc} 
\usepackage[inline]{enumitem}   
\usepackage{graphicx}
\usepackage{multirow}
\usepackage{subfigure}
\usepackage{array}
\usepackage{amssymb}
\usepackage{amsmath}

\DeclareMathOperator*{\argmin}{arg\,min}

\startlocaldefs
\endlocaldefs

\begin{document}
	
	\begin{frontmatter}
		
		\begin{fmbox}
			\dochead{Research}
			
			
			\title{CLeaR: An Adaptive Continual Learning Framework for Regression Tasks}
			
			
			\author[
			addressref={aff1},                   
			corref={aff1},                       
			email={yujiang.he@uni-kassel.de}   
			]{\inits{YH}\fnm{Yujiang} \snm{He}}
			\author[
			addressref={aff1},
			email={bsick@uni-kassel.de}
			]{\inits{BS}\fnm{Bernhard} \snm{Sick}}
			
			
			\address[id=aff1]{
				\orgdiv{Intelligent Embedded Systems (IES) Group},
				\orgname{University of Kassel}, 
				\street{Wilhelmsh\"{o}her Allee 71 - 73},                     %
				\city{Kassel},                              
				\cny{Germany}                                    
			}
			
			
			
			
			
			\begin{abstractbox}
				
				\begin{abstract} 
					Catastrophic forgetting means that a trained neural network model gradually forgets the previously learned tasks when being retrained on new tasks.
					Overcoming the forgetting problem is a major problem in machine learning.
					Numerous continual learning algorithms are very successful in incremental learning of classification tasks, where new samples with their labels appear frequently. 
					However, there is currently no research that addresses the catastrophic forgetting problem in regression tasks as far as we know.
					This problem has emerged as one of the primary constraints in some applications, such as renewable energy forecasts. 
					This article clarifies problem-related definitions and proposes a new methodological framework that can forecast targets and update itself by means of continual learning. 
					The framework consists of forecasting neural networks and buffers, which store newly collected data from a non-stationary data stream in an application. 
					The changed probability distribution of the data stream, which the framework has identified, will be learned sequentially.
					The framework is called CLeaR (\textbf{C}ontinual \textbf{Lea}rning for \textbf{R}egression Tasks), where components can be flexibly customized for a specific application scenario. 
					We design two sets of experiments to evaluate the CLeaR framework concerning fitting error (training), prediction error (test), and forgetting ratio.
					The first one is based on an artificial time series to explore how hyperparameters affect the CLeaR framework.
					The second one is designed with data collected from European wind farms to evaluate the CLeaR framework's performance in a real-world application.
					The experimental results demonstrate that the CLeaR framework can continually acquire knowledge in the data stream and improve the prediction accuracy.
					The article concludes with further research issues arising from requirements to extend the framework.
				\end{abstract}
				
				
				
				
			\end{abstractbox}
		\end{fmbox}
		
	\end{frontmatter}




\section*{Introduction}
In the late 1980s, McCloskey and Cohen~\cite{mccloskey1989catastrophic} and Ratcliff~\cite{ratcliff1990connectionist} observed a phenomenon where the well-learned knowledge of connectionist models is erased by new knowledge under specific conditions when the models learn new tasks successively.
It is referred to as Catastrophic Forgetting or Catastrophic Interference. 
The challenge is supposed to be a general problem existing in different kinds of neural networks, e.g., back-propagation neural networks and unsupervised neural networks.
Each neuron of a layer is connected to all neurons of the next layer in a neural network. 
Weights control the strength of the connection between two neurons.
A weight vector can be expressed as a column vector $\mathbf{w}=(w_{1}, w_{2},..., w_{N})^{\mathrm{T}}$ with $N$ values.
Some weights have a significant influence on more than one task. For example, $w_{1}, w_{2}, w_{3}$ are important for task $1$ and $w_{1}, w_{4}, w_{5}$ are important for task $2$. 
In this case, the overlapped weight $w_{1}$ could be adjusted during learning task $2$ sequentially, which is one of the main reasons for Catastrophic Forgetting.

Overcoming the forgetting problem is a crucial step in implementing real intelligence. 
Models require plasticity for learning and integrating new knowledge as well as stability for consolidating what models have learned previously. 
Excessive plasticity can cause the acquired knowledge to be erased while learning new tasks.
On the other hand, successively learning new tasks can become more challenging due to extreme stability.
It is the so-called stability-plasticity dilemma~\cite{mermillod2013stability}. 

Many researchers have proposed continual learning (CL) algorithms to solve the problem, such as in~\cite{kirkpatrick2017overcoming},~\cite{zenke2017continual}, and~\cite{maltoni2019continuous}. 
A three-way categorization for the most common CL strategies is described in~\cite{maltoni2019continuous}: (1) regularization strategies, (2) rehearsal strategies, and (3) architectural strategies.
Similarly, CL strategies are grouped as (1) prior-focused approaches, (2) likelihood-focused approaches, and (3) dynamic architectures in~\cite{farquhar2019unifying}.
These categorizations standardize terminologies and outline a distinct research direction for the CL community.
CL algorithms have been proven successful in supervised learning and reinforcement learning to train several tasks sequentially without forgetting the acquired ones. 
Application scenarios cover handwriting recognition~\cite{v.2018variational, chaudhry2018riemannian, van2018generative}, image classification~\cite{zenke2017continual}, sequentially learning to play games in a reinforcement learning setting~\cite{kirkpatrick2017overcoming} and much more. 

To our best knowledge, some of the standard CL benchmarks are the reconstructions of well-known datasets, such as permuted MNIST~\cite{goodfellow2013empirical, srivastava2013compete}, where the CL tasks are obtained by scrambling the pixel positions in the MNIST dataset~\cite{lecun-mnisthandwrittendigit-2010}.
Moreover, some datasets are explicitly generated to evaluating CL algorithms, for example, CORe50~\cite{lomonaco2017core50} for continuous object recognition. 

Most of the past research focused mainly on classification tasks rather than regression tasks, where the Catastrophic Forgetting problem usually occurs as well.
For example, establishing regional smart grids requires power generation and consumption forecasts. 
In~\cite{gensler2016solarforecast} and~\cite{he2020forecasting}, neural networks are used to forecast renewable energy generation with weather prediction data.
Note that weather data is non-stationary, where the probability distribution changes over time, e.g., from summer to winter. 
In this situation, training neural networks has to be delayed until sufficient data is collected. 
Otherwise, the networks could be overfitted to the limited training dataset.
Furthermore, unseen situations, e.g., the extreme weather conditions, updating/damaging/ageing of generators, and climate changes, can be called special events. 
The neural networks have to update themselves by continually learning these situations when they appear in the application.
Besides, power consumption regarding a household or a factory is also easily affected by unpredictable things, such as purchasing new equipment or hiring more employees. 
These factors can change the obtained mapping between inputs and outputs.
Under these conditions where historical data may be private, unrecorded, or too cumbersome to be retrained, the trained models have to learn new knowledge and consolidate the previously-stored internal representations without the help of old data.

The \textbf{main contribution} of this article is to propose a framework called CLeaR (\textbf{C}ontinual \textbf{Lea}rning for \textbf{R}egression Tasks) for continually learning the identified changes in non-stationary data streams.
Moreover, the framework is tested in two sets of experiments to assess its performance and analyze its hyperparameters' effects.  
The CLeaR framework consists of neural networks for prediction and buffers for storing new data.
We calculate an error between the prediction and the corresponding true value at each point in time.
The new data is labeled by comparing the error to a dynamically adjustable threshold. 
If the error is larger than the threshold, the data is labeled as a novelty and stored in a finite novelty buffer, or else stored as familiarity in an infinite familiarity buffer.
When the novelty buffer is full, updating will be triggered.
The network will be retrained on the dataset in the novelty buffer using CL. 
The retrained network will then be tested on the familiarity dataset to evaluate how much old knowledge is retained.
After updating, the threshold needs to be re-estimated for the following learning step.
The re-estimation depends on the performances of the updated network on the dataset of both buffers.
Afterwards, the buffers will be emptied.
Updating will be repeated until the novelty buffer is filled again.

The remainder of the article will review the literature regarding CL algorithms and applications. 
Then we will outline the proposed framework's fundamental structure and give an insight into the experimental details. 
Furthermore, we will analyze the experimental results.
This article will conclude with our findings and provide an outlook for future research.

\subsection*{\textbf{Related work}}
This section will start with a brief overview of the recent academic literature regarding approaches and experimental setups designed for CL classification applications.

The changes in data or goals can be defined as new tasks in the CL community.
For example, a model is expected to learn new instances of the same class while retaining its knowledge about the previous instances, or to learn new instances of different classes without losing accuracy on previous classes, or to learn new instances of the known and unknown classes.
These are defined as different CL scenarios in ~\cite{lomonaco2017core50}.
In both~\cite{maltoni2019continuous} and~\cite{farquhar2019unifying}, CL algorithms are categorized into three groups in a similar way: 
\begin{itemize}
    \item{\textit{Prior-focused approaches}} denote that the posterior probability of $N$ tasks is a product of the likelihood of the $N$th task and the posterior probability of the first $N-1$ tasks, as
    \begin{eqnarray}
    \centering
    \begin{aligned}
    \resizebox{.8\hsize}{!}{$P\left(\Theta|D_{1},~\dots~D_{N}\right)=\frac{P\left(D_{N}|\Theta\right)P\left(\Theta|D_{1},~\dots~D_{N-1}\right)}{P\left(D_{N}|D_{1},~\dots~D_{N-1}\right )}$}.
    \end{aligned}
    \label{formula:prior_focused}
    \end{eqnarray}
    As a regularization, the posterior probability of the first $N-1$ tasks is added in the loss function to avoid changing the weights that are important for previous tasks.
    Well-known prior-focused algorithms include Elastic Weight Consolidation (EWC)~\cite{kirkpatrick2017overcoming}, Synaptic Intelligence (SI)~\cite{zenke2017continual}, Variational Continual Learning~\cite{v.2018variational}. 
    In~\cite{chaudhry2018riemannian}, a generalization of EWC++ and SI was proposed, which is referred to as the RWalk algorithm.
    
    \item {\textit{Likelihood-focused approaches}} require a subset of randomly selected samples from the original dataset of the previous $N-1$ tasks, or a dataset generated by a generative network that has learned the tasks, see in~\cite{shin2017continual},~\cite{farquhar2019unifying}, and~\cite{van2018generative}.
    
    \item {\textit{Dynamic architectures}} enable neural networks to learn CL tasks sequentially by adjusting the networks' architecture for specific applications. 
    Progressive Networks, Learning Without Forgetting (LWF), and Less-Forgetting Learning have been introduced in~\cite{rusu2016progressive, li2017learning, jung2016less}, respectively.
\end{itemize}

The above algorithms have been evaluated in multi-task scenarios with the reshaped versions of famous datasets, e.g., MNIST and CIFAR-10/CIFAR-100. 
In these scenarios, a model learns a new, isolated task in a sequence while remembering how to solve the learned tasks. 
However, there are no class overlaps among the different tasks.
For example, in~\cite{zenke2017continual} the MNIST dataset is split into five tasks, one of which contains two labels (two digits).
The model can classify data to the correct group only if the information regarding the current task is given. 
In this case, the model learns how to solve a series of discrete tasks rather than keep learning knowledge to address incremental problems. 
The experimental setups and the datasets do not allow for a fair comparison among the CL algorithms. 
Lomonaco et al.~\cite{lomonaco2017core50} create CORe50 specifically for single-incremental-task scenarios, which can be seen as a test benchmark for continuous object recognition. Similarly, the iCubWorld benchmark~\cite{pasquale2016object} is designed for robotic vision challenges, where comparison among various CL approaches is feasible.

Besides, continual learning should be considered in regression as well.
In~\cite{he2020continuous}, He et al. propose two CL application scenarios for establishing regional smart grids: the task-domain incremental scenario and the data-domain incremental scenario.
The scenarios are applicable for forecasting power, including renewable energy generation and power consumption in the middle-/low-voltage grid.
Moreover, performances of four CL algorithms (EWC, Online-EWC, SI, and LWF) are evaluated concerning accuracy, forgetting ratio, and training time in the two scenarios. 
However, prior knowledge about new tasks is given in their experimental setup, which means that models know when new tasks will occur without novelty detection. 
Therefore, this setup is incompatible with the real world.

In~\cite{farquhar2018towards}, Farquhar et al. conclude that an inappropriate experimental design could misrepresent the performances of the well-known CL approaches. 
Therefore, they suggest five requirements for evaluating CL algorithms and demonstrate their necessities. 
The five requirements are: (1) cross-task resemblances; (2) shared output head; (3) no test-time assumed task labels; (4) no unconstrained retraining on old tasks; and (5) more than two tasks.

These suggestions are worth being considered in our experimental setup and inspire us to design the CLeaR framework.
(1) Most novelties are due to changes of data $P\left(X\right)$ or targets $P\left(Y|X, \Theta\right)$.
The dataset of every full novelty buffer can be viewed as a new task that resembles the previous tasks.
(2) The neural network outputs the power value prediction, and the new tasks will not require a change of the network's architecture.
(3) The prior knowledge regarding the new task, such as when the new task appears or what the distribution of the new task is, is unknown in the application. 
Updating is triggered automatically only when the finite novelty buffer is filled in our experimental setup. 
(4) Considering that privacy laws might prohibit the long-term storage of historical datasets, we retrain the neural network only on the dataset newly collected in applications and delete it after updating.
(5) More tasks will appear as the probability distribution or the mapping between inputs and outputs changes over time.
\section*{Power forecasts using deep neural networks}
At the beginning of this section, we list the chosen mathematical notations in Table~\ref{tab:notations}.
It can help readers to understand the following mathematical expressions.
Besides, we use superscript $^{\mathrm{T}}$ to denote the transpose of a matrix or a vector and $T$ to denote the number of tasks.
\begin{table}[t]
	\caption{Notations.}
	\begin{tabular}{c@{\hskip 0.1mm}c}
		\hline
        Symbol & Definition \\ 
        \hline
        $X$ & The $N \times M$ matrix with $N$ feature samples\\
        $Y$ & The matrix with $N$ measurements associated with $X$\\
        $\hat{Y}$ & The matrix with $N$ predictions associated with $X$\\
        $\mathbf{x}_n$ & The $n$th column feature vector, $\mathbf{x}_n^{\mathrm{T}}=X_{n,:}$\\
        $y_n$ & The $n$th measurement, $y_n = Y_{n,:}$\\
        $\hat{y}_n$ & The $n$th prediction, $\hat{y}_n = \hat{Y}_{n,:}$\\
        $D$ & The dataset, $D=\{\left ( \mathbf{x}_n, y_n \right ) |  n=1, \dots N \}$\\
        $\Theta_{l}$ & The weight matrix of the $l$th layer, $\Theta_{l} \in \mathbb{R}^{{l-1} \times {l}}$\\
        $\theta^{i}_{l}$ & The $i$th element of $\Theta_{l}$\\
        $f_l$ & The activation function of the $l$th layer, $f_l\colon \mathbb{R} \to \mathbb{R}$\\
        $f_{\Theta}\left( \cdot \right)$ & The neural network with given weight martrix $\Theta$\\
        $\textbf{z}_l$ & The output column vector of the $l$th layer\\
        $L$ & Loss function \\
        $P$ & Probability density \\
        $\mathcal{N}(\mu, \sigma^2)$ & Gaussian distribution with mean $\mu$ and variance $\sigma^2$\\
        $\mathbf{F}_{t}$ & The Fisher information matrix of the $t$th task\\ 
        $F_{t}^{i}$ & The $i$th diagonal element of $\mathbf{F}_{t}$\\
        \hline
	\end{tabular}
    \label{tab:notations} 
\end{table}

Deep neural networks are a kind of machine learning inspired by biological neural networks to model nonlinear dependencies in high dimensional data.
Compared with traditional high dimensional data reduction techniques, such as principal component analysis (PCA), the multiple deep layers of a neural network can extract representations efficiently from massive data to provide predictive performance gains. 

The rest of this section will start with formulating the problem. 
Moreover, this section will illustrate the probability distribution changes in the experimental dataset over time and clarify how the CLeaR framework works in a general power forecasting workflow.

\subsection*{\textbf{Problem formulation}}
Deep neural networks can output a prediction $y_{n}$ with given a high-dimensional input $\textbf{x}_n$. 
The goal of training is to find a mapping between $\textbf{x}_n$ and $y_n$. 
A general deep neural network consists of a series of hidden layers, which can be formulated as: 
\begin{eqnarray}
\centering
\small
\begin{aligned}
\textbf{z}_{l} = f_l\left(\Theta_{l}^{\mathrm{T}} \textbf{z}_{l-1}\right)
\end{aligned}
\label{formula:hidden_layer}
\end{eqnarray}
with an output column vector $\textbf{z}_{l-1}$ of the $l-1$th layer, where $\textbf{z}_{0}=\textbf{x}_n$. 
Here $\Theta_{l}^{\mathrm{T}}$ denotes a transposed weight matrix, whose dimension is the dimension of the $l-1$th layer by the dimension of the $l$th layer.
A prediction $\hat{y}_n$ of the deep neural network is then
\begin{eqnarray}
\centering
\small
\begin{aligned}
\hat{y}_n &= f_{\Theta}\left(\textbf{x}_n\right) \\&= f_L\left (\Theta_L^T f_{L-1}\left(\dots \Theta_2^T f_{1}\left(\Theta_1^T \textbf{x}_n \right)\right)\right),
\end{aligned}
\label{formula:dnn}
\end{eqnarray}
where $L$ is the number of layers.

Training a neural network is to minimize the defined loss function.
In this section, we explain the process with an example of Mean Square Error (MSE), i.e.,
\begin{eqnarray}
\centering
\small
\begin{aligned}
\L\left(Y,\hat{Y}\right) =\frac{1}{N}\sum_{n=1}^{N}\| y_n - \hat{y}_n\|^2_{2}.
\end{aligned}
\label{formula:mse_loss}
\end{eqnarray}
In order to avoid overfitting, a regularization term is usually added in Eq.~\ref{formula:mse_loss}, as
\begin{eqnarray}
\centering
\small
\begin{aligned}
L\left(Y,\hat{Y}\right) =\frac{1}{N}\sum_{n=1}^{N}\| y_n - \hat{y}_n\|^2_{2} + \lambda R\left(\Theta\right),
\end{aligned}
\label{formula:mse_loss_r}
\end{eqnarray}
where $\lambda \in \left(0, \infty \right )$ is a hyperparameter that weights the contribution of penalty term $R\left(\Theta\right)$.
Different choices for $R\left(\Theta\right)$ can result in different solutions.

From a probabilistic perspective, according to chapter $9$ in~\cite{deisenroth2020mathematics}, we can assume that we are given an input $\textbf{x}_n$ and the corresponding noisy observation $y_{n} = \hat{y}_n + \epsilon$.
More specifically, we assume further that this noise $\epsilon$ follows an independent and identical Gaussian distribution with zero mean and variance $\sigma^2$.
Therefore, the regression problem can be considered with a likelihood function:
\begin{eqnarray}
\centering
\small
\begin{aligned}
P\left (y_n|\textbf{x}_n, \Theta \right ) = \mathcal{N} \left (f_{\Theta}(\textbf{x}_n), \sigma^2 \right ).
\end{aligned}
\label{formula:likelihood}
\end{eqnarray}
When we are given the datasets $X$ and $Y$, Eq.~\ref{formula:likelihood} can be expressed as
\begin{eqnarray}
\centering
\small
\begin{aligned}
P\left (Y|X, \Theta \right ) &= \prod_{n=1}^{N} P \left ( y_{n}|\textbf{x}_n, \Theta \right )\\
                        &= \prod_{n=1}^{N} \mathcal{N} \left (f_{\Theta}(\textbf{x}_n), \sigma^2 \right ),
\end{aligned}
\label{formula:likelihood_2}
\end{eqnarray}
where assumed that the $y_i$ and $y_j$ are conditionally independent given their feature vectors $\textbf{x}_i$ and $\textbf{x}_j$.
To avoid overfitting during training, we seek parameters $\Theta$ that maximize the posterior distribution $P\left(\Theta|X, Y\right)$ instead of the likelihood.
We can obtain the posterior distribution by applying Bayes' theorem as
\begin{eqnarray}
\centering
\small
\begin{aligned}
P\left ( \Theta|X, Y \right ) = \frac{P \left ( Y|X,\Theta \right)P\left( \Theta \right)}{P\left( Y|X \right)}
\end{aligned}
\label{formula:posterior_bayes}
\end{eqnarray}
Note that the posterior distribution depends on the given $X$ and $Y$. 
If the statistical properties of the distributions change, such as the mean or the variance, new parameter values will become optimal.

\subsection*{\textbf{Change in probability distribution}}
\begin{figure*}[t]
	\subfigure[From 1 to 2500]{
		\includegraphics[width=.1735\textwidth]{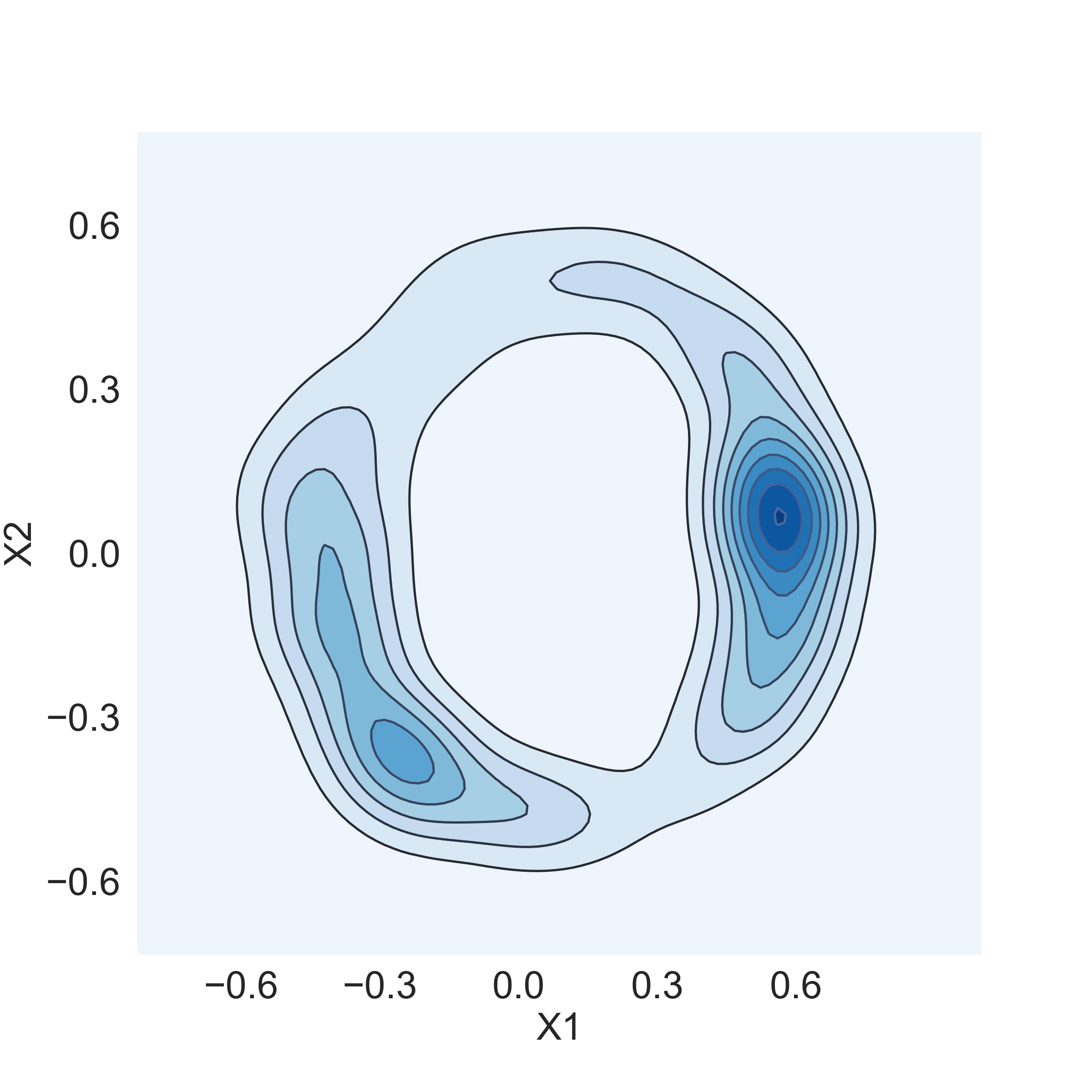}
		\label{subfig:weather_0}
	}\hfill
	\subfigure[From 2501 to 5000]{
		\includegraphics[width=.1735\textwidth]{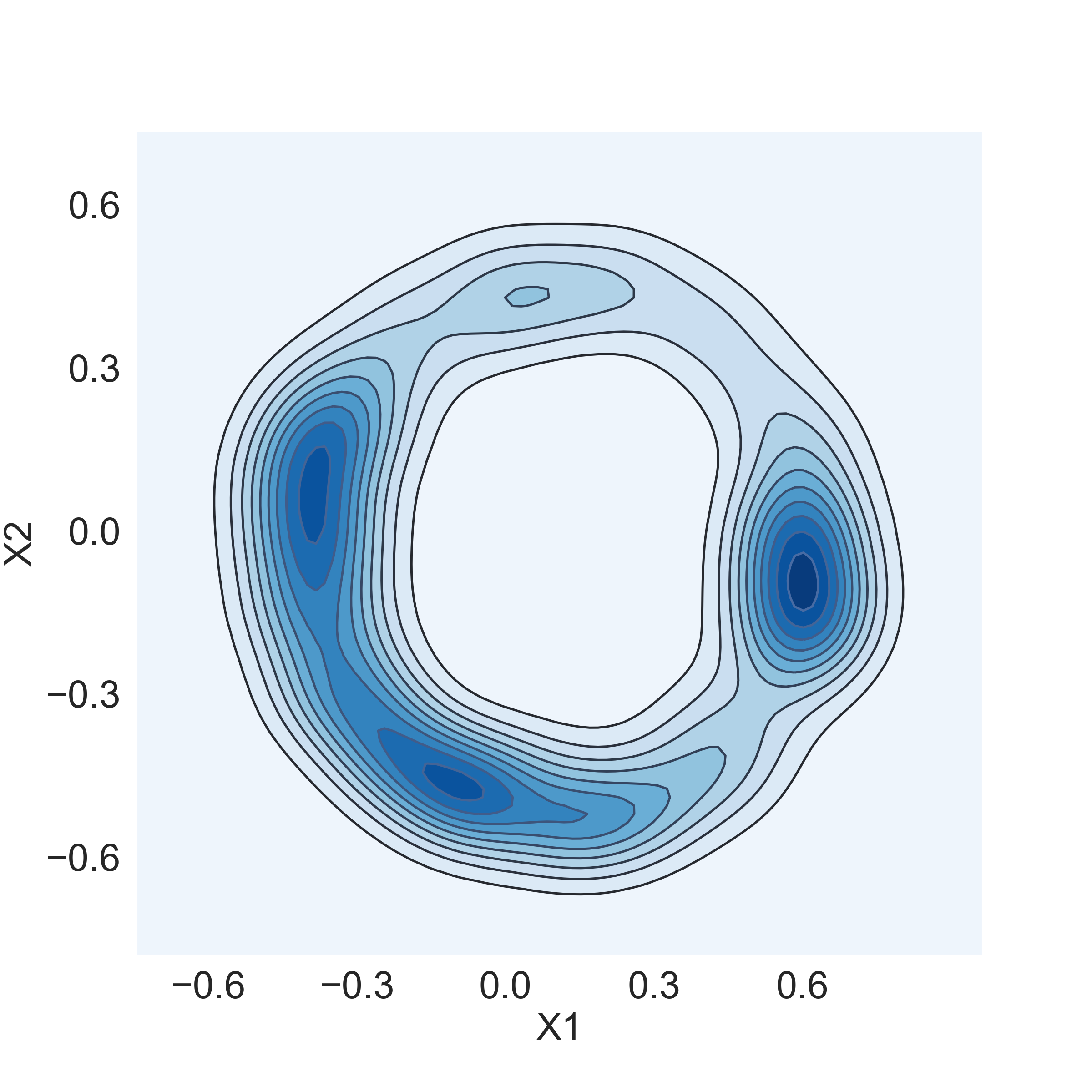}
		\label{subfig:weather_1}
	}\hfill
	\subfigure[From 5001 to 7500]{
		\includegraphics[width=.1735\textwidth]{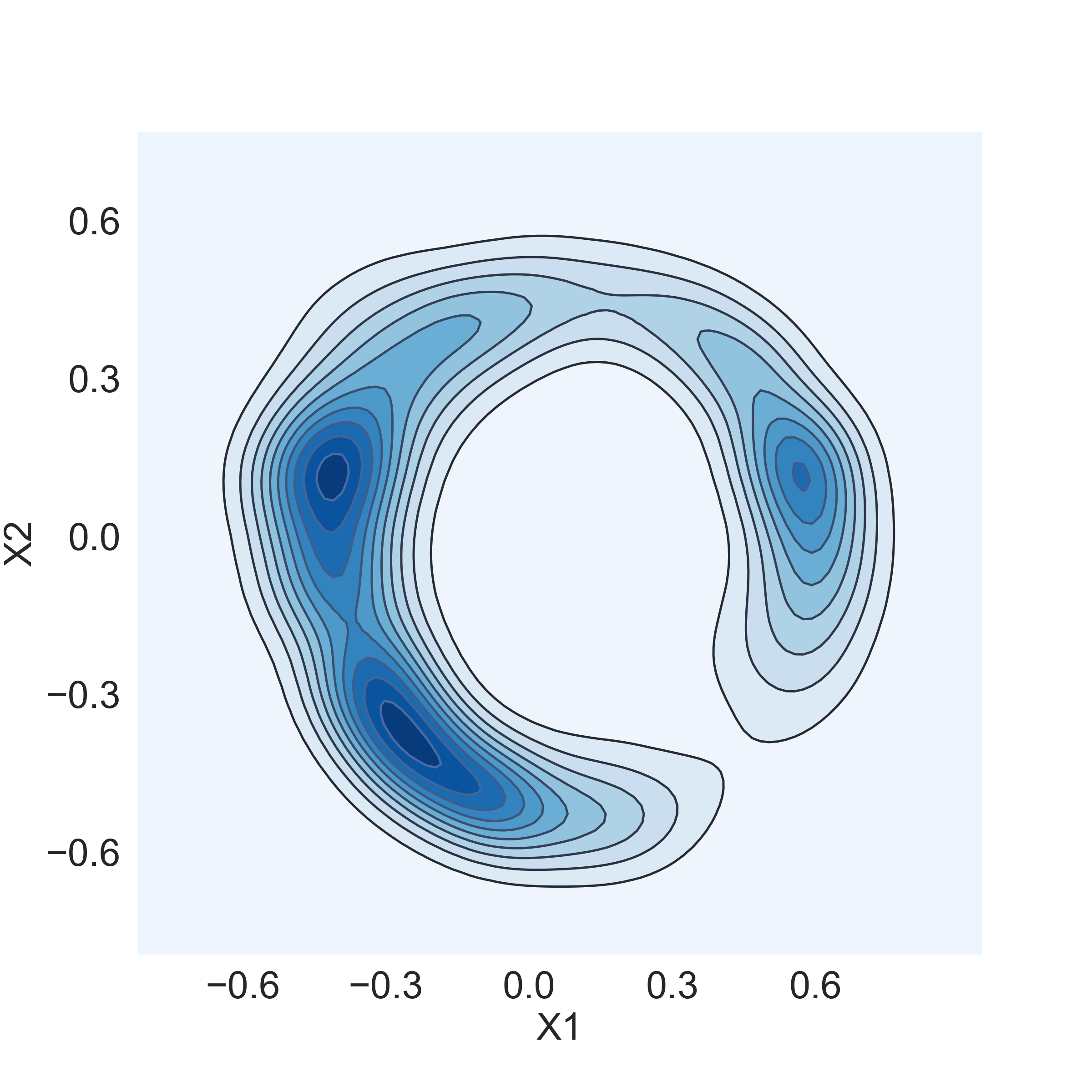}
		\label{subfig:weather_2}
	}\hfill
	\subfigure[From 7501 to 10000]{
		\includegraphics[width=.1735\textwidth]{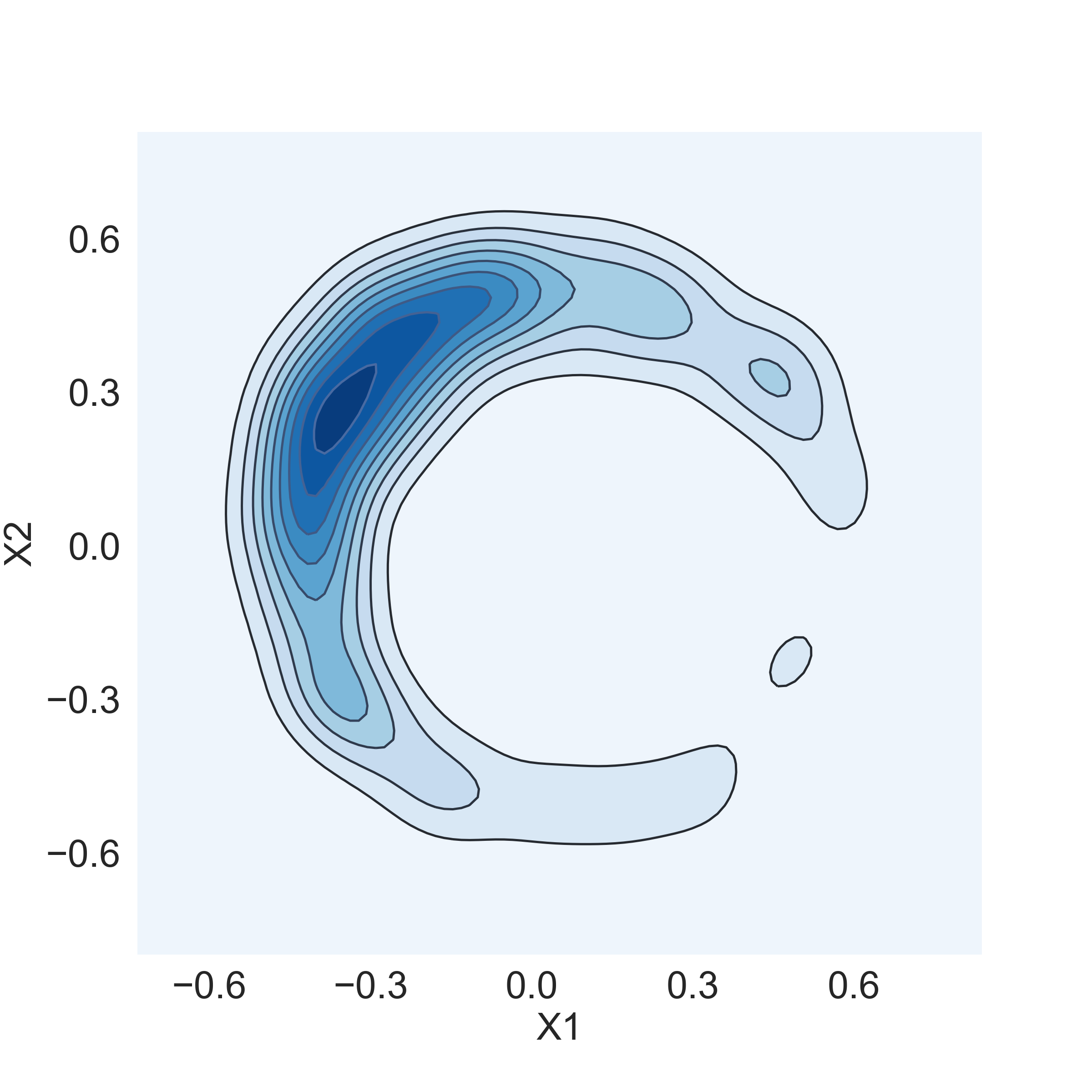}
		\label{subfig:weather_3}
	}\hfill
	\subfigure[10000 samples]{
		\includegraphics[width=.1735\textwidth]{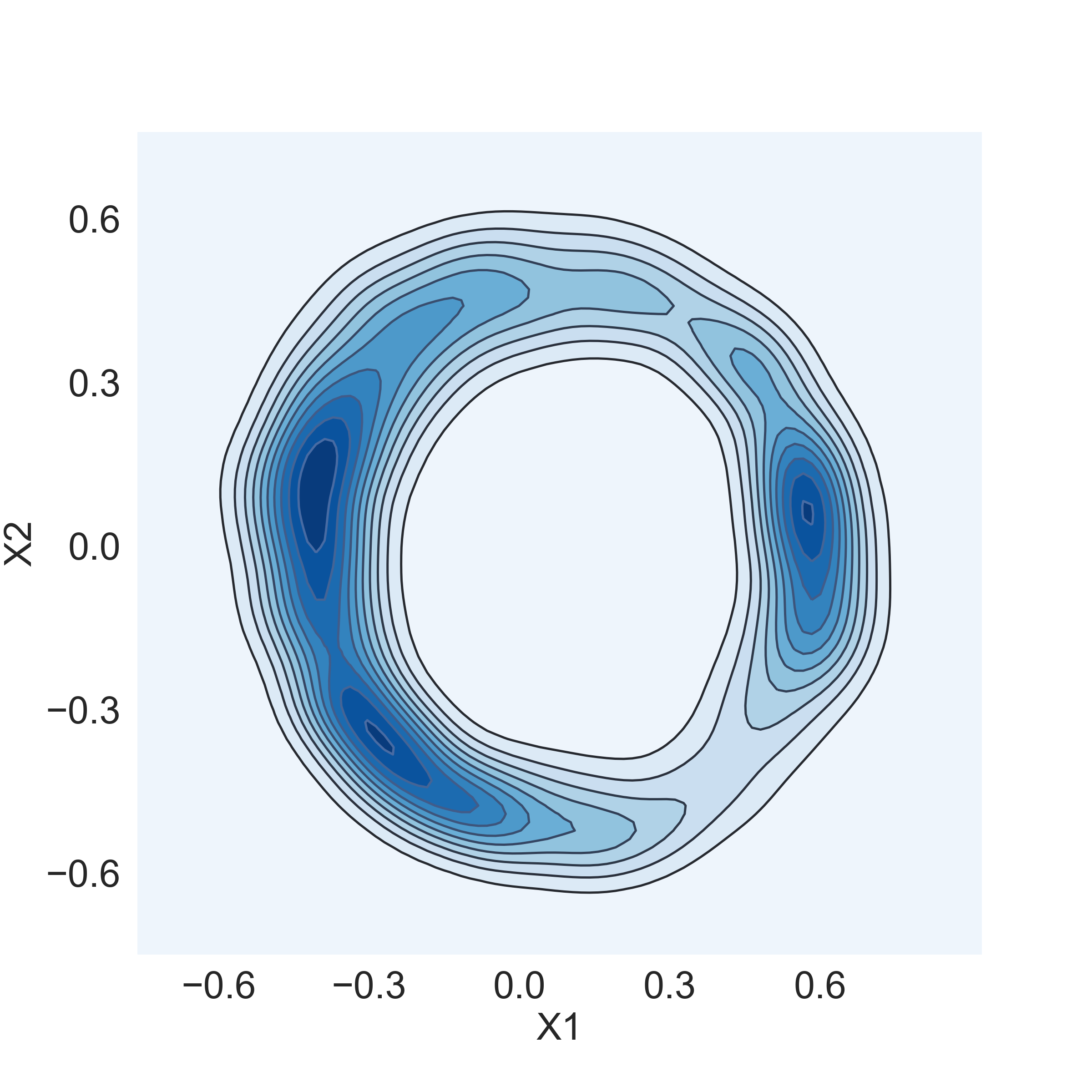}
		\label{subfig:weather_all}
	}
	\caption{\csentence{Distributions of the inputs change over time.} We split 10000 samples into four sections sequentially over time, each of which has 2500 samples, and visualize their two-dimensional distributions from (a) to (d). (e) illustrates the distribution of the same 10000 samples. Both axes in the sub-figures, X1 and X2, indicate the first two principal components extracted from seven-dimensional weather features.}
	\label{fig:distribution_weather}
\end{figure*}

\begin{figure*}[t]
	\subfigure[From 1 to 2500]{
		\includegraphics[width=.1735\textwidth]{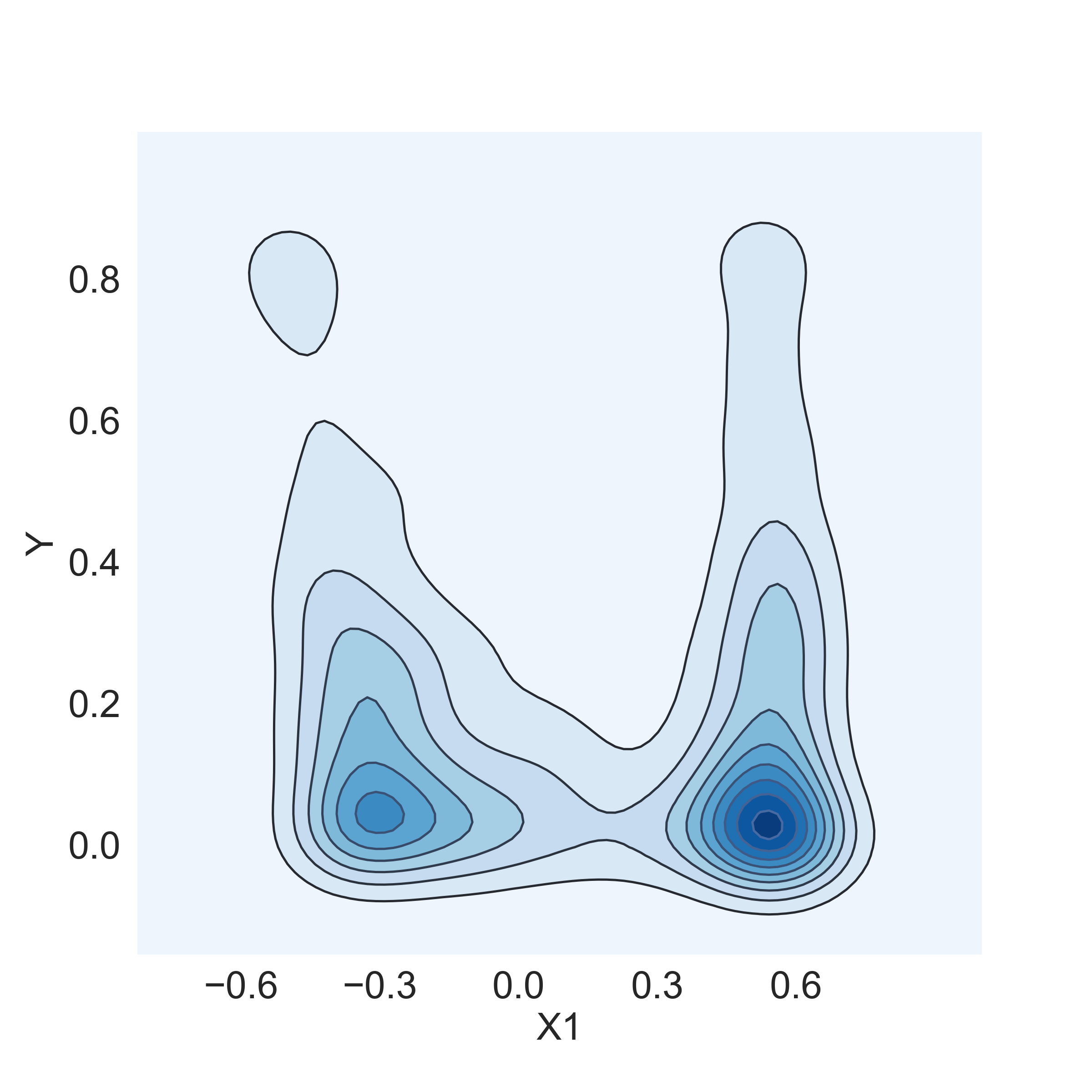}
		\label{subfig:power_0}
	}\hfill
	\subfigure[From 2501 to 5000]{
		\includegraphics[width=.1735\textwidth]{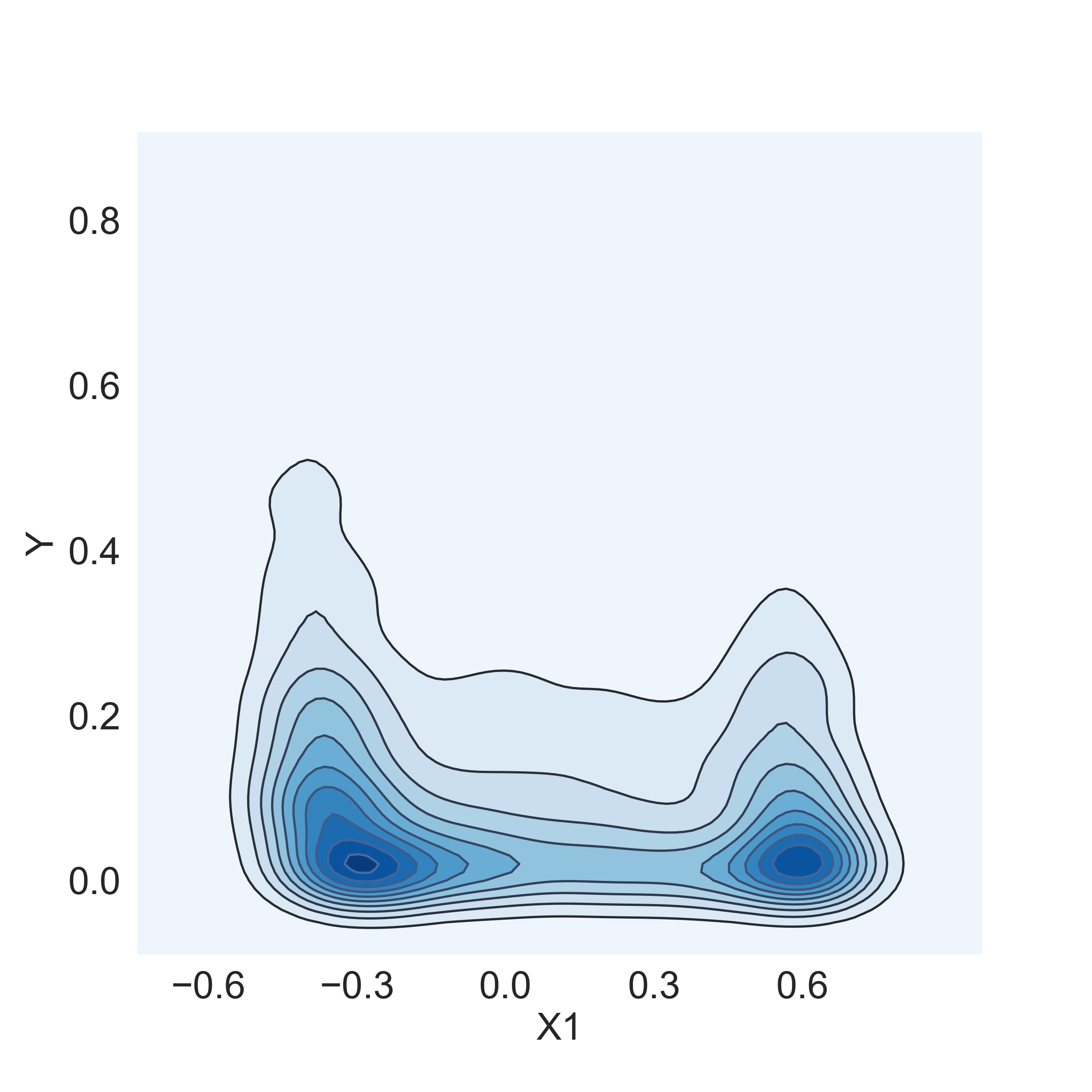}
		\label{subfig:power_1}
	}\hfill
	\subfigure[From 5001 to 7500]{
		\includegraphics[width=.1735\textwidth]{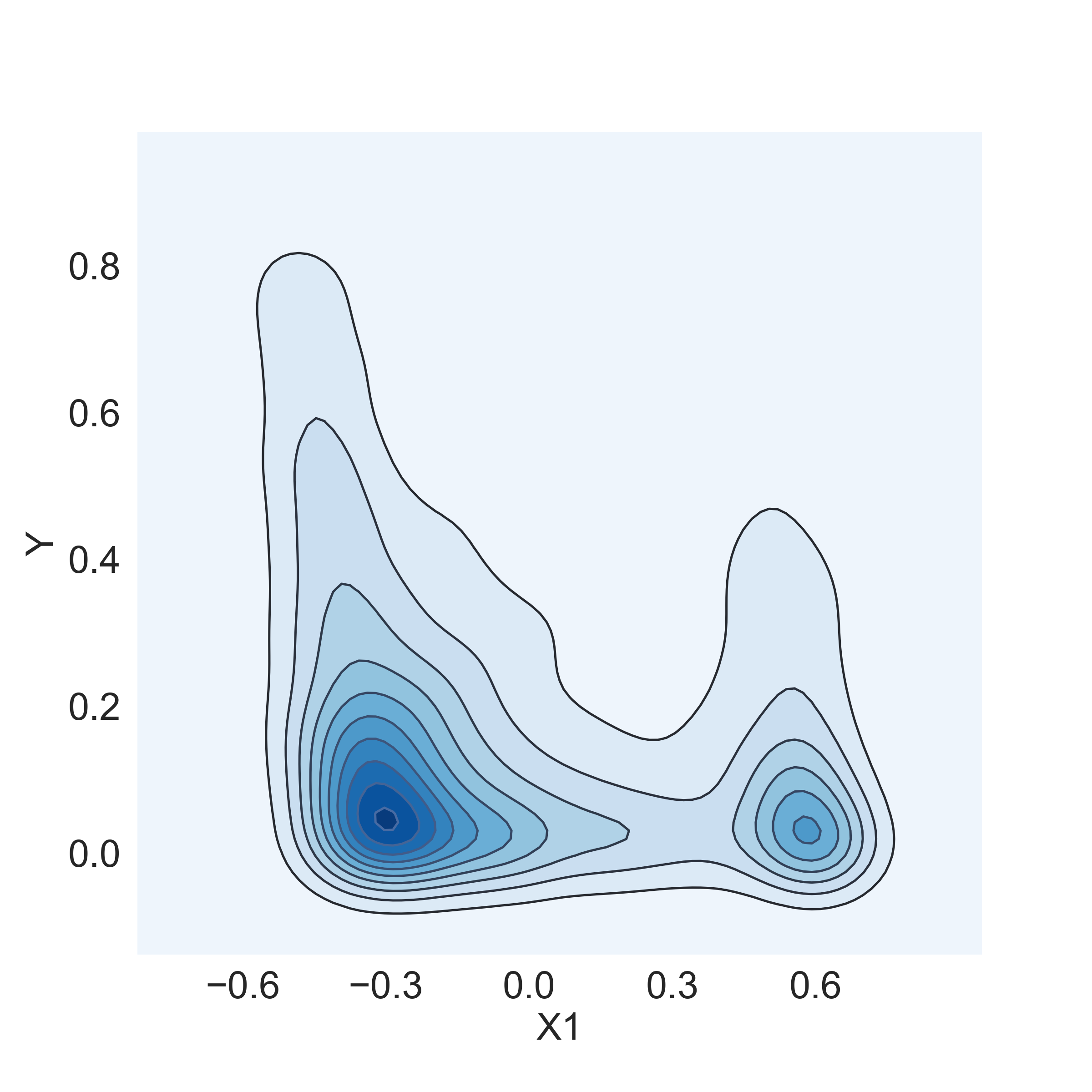}
		\label{subfig:power_2}
	}\hfill
	\subfigure[From 7501 to 10000]{
		\includegraphics[width=.1735\textwidth]{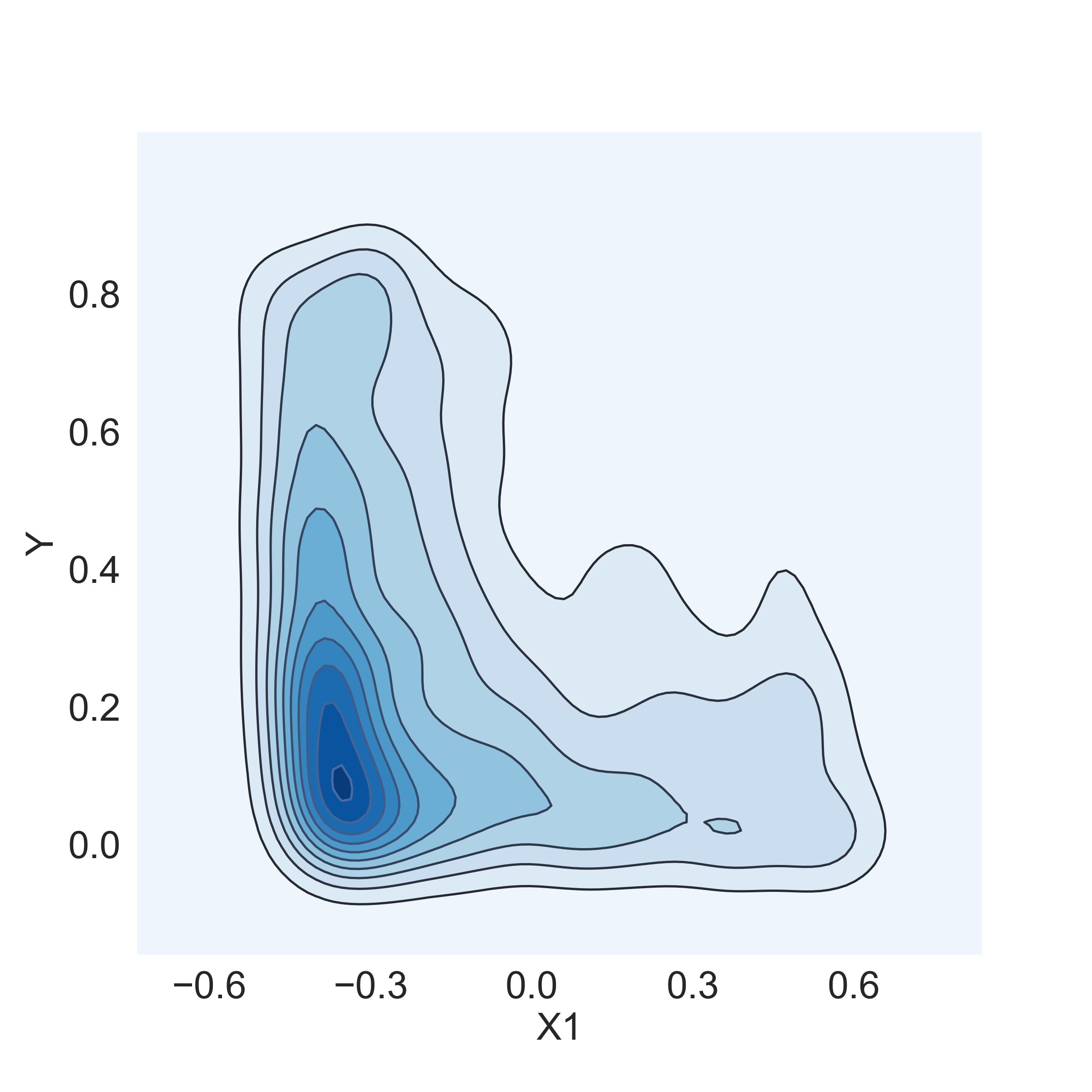}
		\label{subfig:power_3}
	}\hfill
	\subfigure[10000 samples]{
		\includegraphics[width=.1735\textwidth]{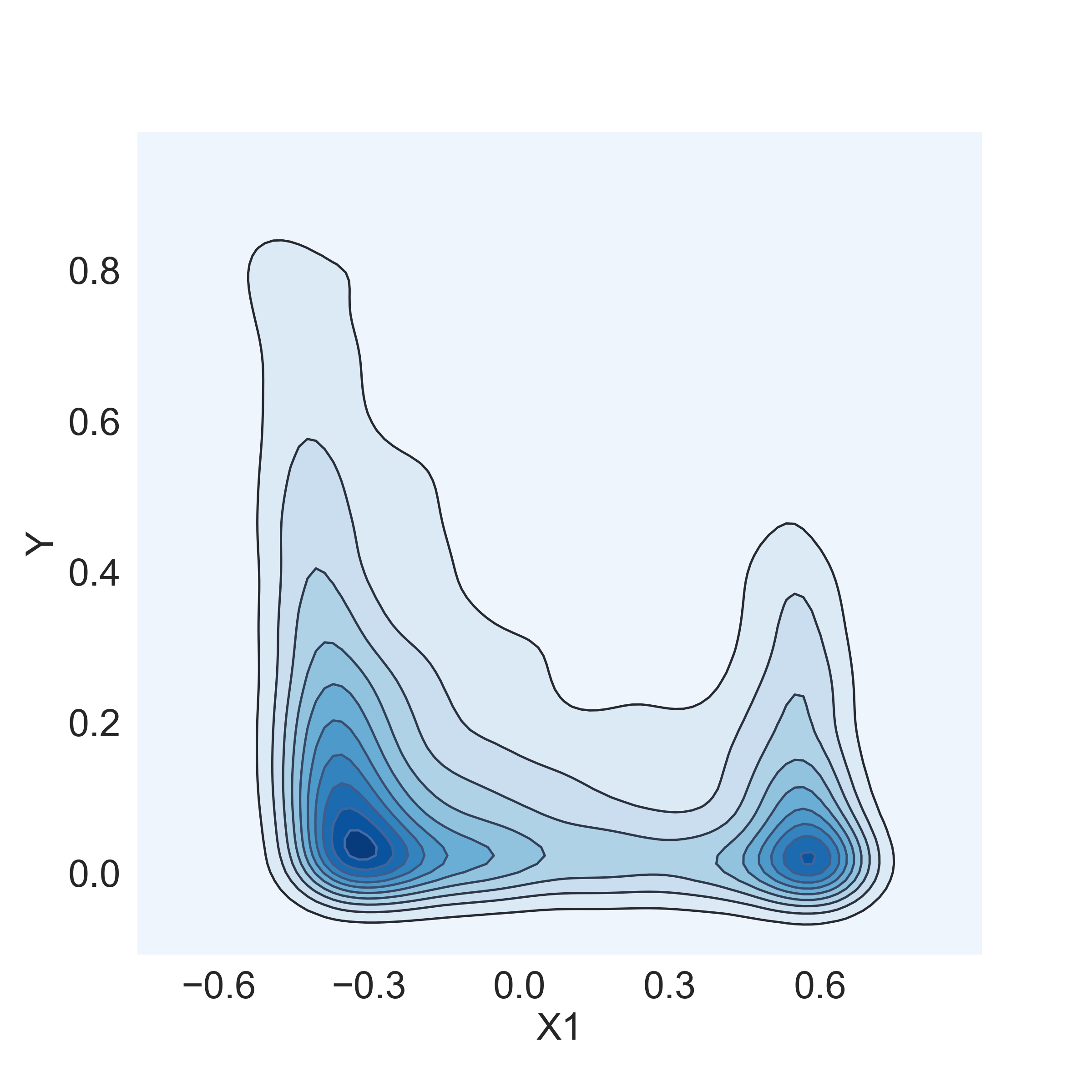}
		\label{subfig:power_all}
	}
	\caption{\csentence{Distributions of the input and the output change over time.} We split 10000 samples into four sections sequentially over time, each of which has 2500 samples, and visualize their two-dimensional distributions from (a) to (d). (e) illustrates the distribution of the same 10000 samples. Both axes in the sub-figures, X1 and Y, indicate the first principal component extracted from seven-dimensional weather features using PCA and the power value, respectively.}
	\label{fig:distribution_power}
\end{figure*}
A learning task can be described as changes in distributions of data $P\left(X\right)$ or targets $P\left(Y|X, \Theta\right)$.
In our experimental setup, meteorological features and the power measurements are viewed as the inputs and outputs of the neural network, respectively. 
Some meteorological features, e.g., temperature and wind direction, fluctuate yearly periodically. 
Also, renewable energy generation depends on meteorological conditions. 
Therefore, the joint probability distribution $P\left(X, Y\right)$ changes period by period.
Mathematically, the change can be expressed as follows:
\begin{eqnarray}
\centering
\small
\exists t_0 \neq t_1: P\left(X_{t_{0}},Y_{t_{0}}\right) \neq P\left(X_{t_{1}},Y_{t_{1}}\right),
\label{formula:inequality}
\end{eqnarray}
where $X$ and $Y$ represent a batch of features and power measurements respectively, which are measured from a non-stationary data stream.
If two measurement periods $t_0$ and $t_1$ are far apart, both distributions are different due to concept drift.
The formula:
\begin{eqnarray}
\centering
\small
P\left(X,Y\right) = P\left(Y|X\right)P\left(X\right)
\label{formula:conditional_p}
\end{eqnarray}
demonstrates that $P\left(X,Y\right)$ is affected by the change in the probability distribution of the inputs and the obtained mapping.

Figs.~\ref{fig:distribution_weather} and~\ref{fig:distribution_power} illustrate the change regarding $P\left(X\right)$ and $P\left(Y|X\right)$ based on data from a European wind farm dataset. 
A sample of the dataset contains seven-dimensional meteorological features and a scalar power value.
The first two principal components are extracted from the seven features using PCA and labeled as $X1$ and $X2$.
$64.19\%$ of the overall variance is explained by the two components. 
We split 10000 samples from the dataset into four sections sequentially over time, each of which has 2500 samples. 
The distribution of each section regarding $X1$ and $X2$ is plotted in Fig.~\ref{fig:distribution_weather}.
Similarly, the distributions of four sections regarding $X1$ and the power $Y$ are shown in Fig.~\ref{fig:distribution_power}.
The distributions of $P\left(X1, X2\right)$ and $P\left(X1, Y\right)$ with the same 10000 samples are plotted in Fig.~\ref{subfig:weather_all} and Fig.~\ref{subfig:power_all} in comparison to other sub-figures.

The two-dimensional probability distribution of $P\left(X1, X2\right)$ with 10000 samples shows a circular shape, as shown in Fig.~\ref{subfig:weather_all}. 
From Figs.~\ref{subfig:weather_0} to~\ref{subfig:weather_3}, the center of the contour line regarding each section moves clockwise along the circle.
This movement illustrates the periodic change in the $P\left(X\right)$ over time. 
Similarly, we can observe that the center of the distribution $P\left(Y|X1\right)$ moves from right to left over time, as shown in Fig.~\ref{fig:distribution_power}.

Combined with Eq.~\ref{formula:inequality} and the observations, we conclude that a periodic change exists in the dataset.
The model needs to be updated in applications if it was pre-trained only on a limited dataset.

\subsection*{\textbf{Power forecasting workflow}}
\begin{figure*}[t]
	\centering
	\includegraphics[width=0.95\textwidth]{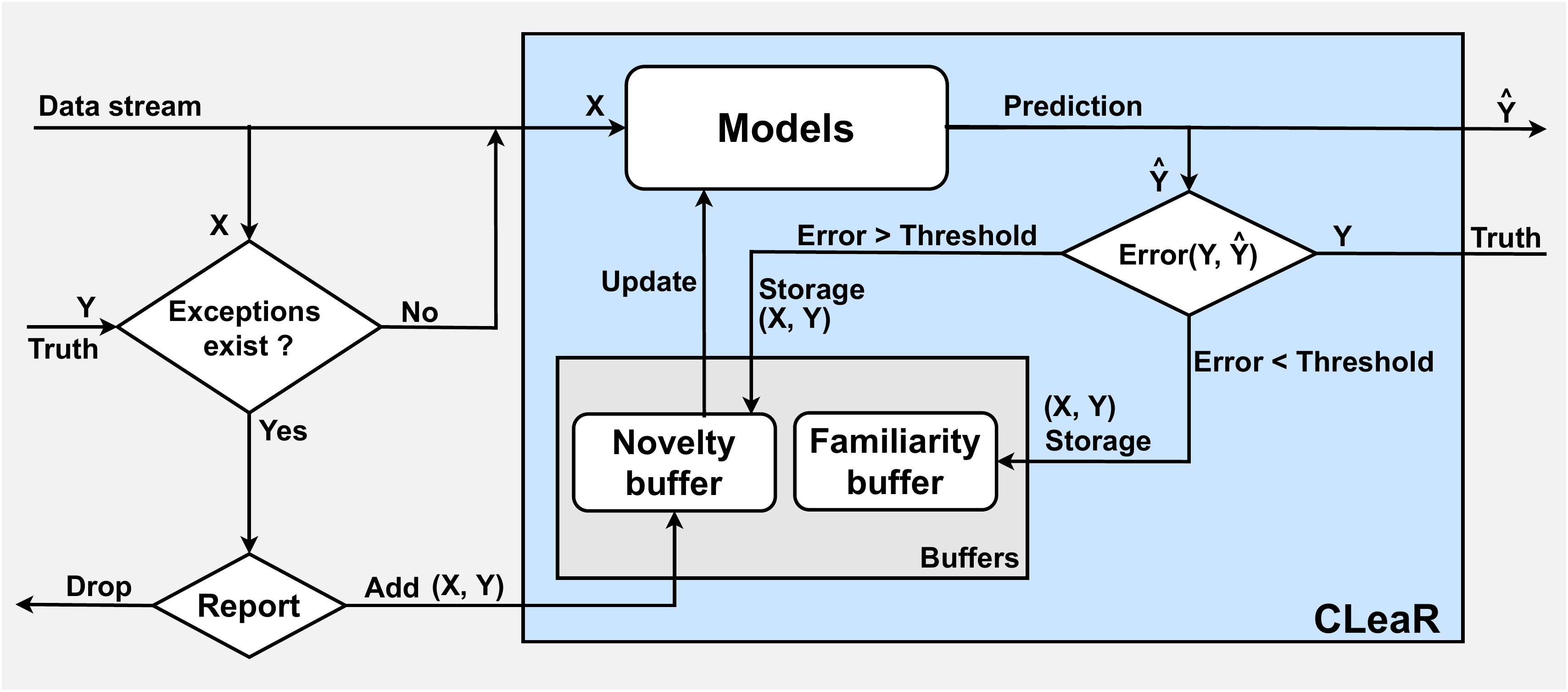}
	\caption{A general workflow of power forecasts comprises four phases: (1) reporting exceptions, (2) predicting targets, (3) storing data, and (4) updating models.}
	\label{fig:workflow}
\end{figure*}
We suggest that a general power forecasting workflow should comprise four phases: (1) reporting exceptions, (2) predicting targets, (3) storing data, and (4) updating models, as shown in Fig.~\ref{fig:workflow}. 

We define that an exception deviating from the expected model is an outcome from an unknown process. 
For example, wind power generators are automatically shut down for protection under extreme weather conditions, such as typhoons or storms. 
Exceptions have to be identified first as the data arrives. 
If it exists in the data stream, the exceptions need to be reported and processed manually. 
Although the dataset used here has been cleaned in the preprocessing phase, these exceptions might appear in applications. 
Labeling and learning exceptions are one of the research challenges in the field of active learning, which is beyond the scope of this article. 
It can be further researched in the future.

The CLeaR framework contains the models for prediction and the buffers for storage.
Once the measurements $Y$ are available, the corresponding data is labeled and stored into the two buffers depending on the preset threshold and the error.
We choose the MSE here for supervised regression tasks, but the method can be customized in other scenarios, e.g., probabilistic forecasts.
The data in the novelty buffer covers the change of probability distribution detected in the data stream.
It is used for retraining the models when the update is triggered.
The data in the familiarity buffer has information that the models are familiar with. 
It can be used for testing whether the models still retain the old knowledge after updating. 
Updating models can be considered as accumulations of knowledge for improving prediction accuracy.
\section*{CLeaR}
\begin{figure*}[t]
	\centering
	\includegraphics[width=0.95\textwidth]{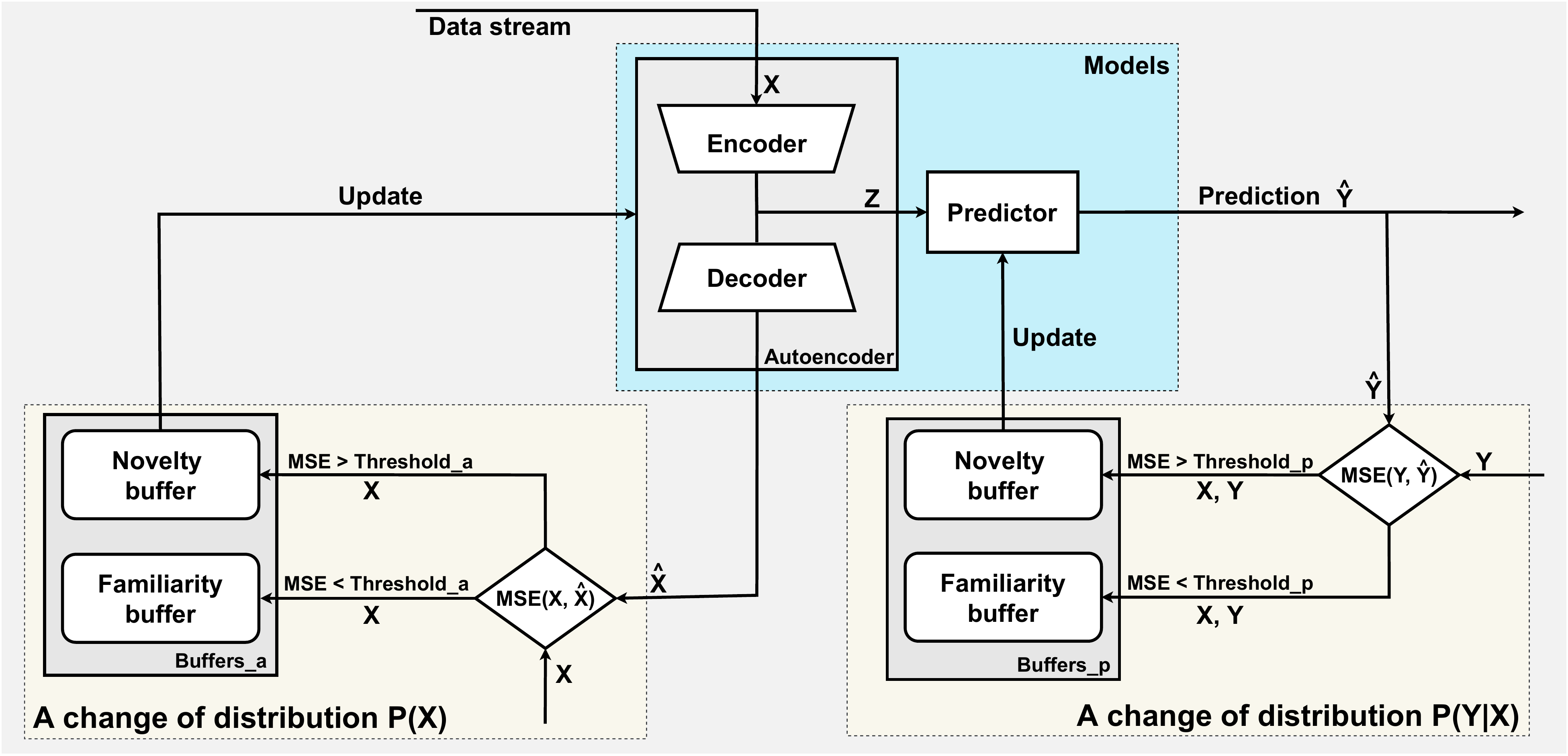}
	\caption{The CLeaR instance contains an autoencoder and a fully-connected neural network in the block \textbf{Models}. The MSE($X$, $\hat{X}$) is compared to \textbf{Threshold\_a} to detect the change of $P(X)$, and the MSE($Y$, $\hat{Y}$) is compared to \textbf{Threshold\_p} to detect the change of $P(Y|X)$. The novelty buffer in \textbf{Buffers\_a} is for updating the autoencoder and the novelty buffer in \textbf{Buffers\_p} is for updating the predictor. Models are evaluated with the data in the familiarity buffer.}
	\label{fig:details}
\end{figure*}
The details of the CLeaR framework will be explained through the instance used in our experiments, as shown in Fig.~\ref{fig:details}. 
In this instance, the block \textbf{Models} contains an autoencoder and a fully-connected neural network for detecting the changes of the distributions $P\left(X\right)$ and $P\left(Y|X\right)$ and for forecasting power values. 
Data is labeled as novelty or familiarity by comparing the MSE to the threshold.
Besides, we adopt Online-EWC to update the models and adjust the threshold dynamically after each update.
We suggest that the components of the CLeaR framework should be selected flexibly depending on the specific application scenario. 

\subsection*{\textbf{Models}}
An autoencoder is a neural network that usually consists of two symmetric parts with a bottleneck between them.
In an undercomplete autoencoder, the bottleneck has a smaller dimension than the input layer, which helps to extract latent representations $\textbf{z}$ from the input.
An autoencoder can reconstruct the input at the output, rather than simply copy the input~\cite{Goodfellow-et-al-2016}. 
The encoder and the decoder can be formulated $\textbf{z}_n=f_{\Theta}\left(\textbf{x}_n\right)$ and $\hat{\textbf{x}}_n=g_{\Phi}\left(\textbf{z}_n\right)$, where $\Theta$ and $\Phi$ are the parameter matrices.
The optimization goal is to minimize the loss function,
\begin{eqnarray}
\centering
\small
\begin{aligned}
\mathit{L}\left(\textbf{x}_n, \Phi, \Theta\right) = \mathit{MSE}\left(\textbf{x}_n, g_{\Phi}\left(f_{\Theta}\left(\textbf{x}_n\right)\right)\right),
\end{aligned}
\label{formula:autoencoder}
\end{eqnarray}
by penalizing the reconstruction being different from the input.
The change of distribution $P\left(X\right)$, as shown in Fig.~\ref{fig:distribution_weather}, can be detected by the reconstruction error of the autoencoder.

At the next step, the extracted representation $\textbf{z}_n$ is fed into the predictor. 
As explained in Eqs.~\ref{formula:mse_loss} and~\ref{formula:mse_loss_r}, optimizing the network in the general supervised setting is to minimize the MSE. 
Thus a true measurement $y_n$ is required.
The predictor, a fully-connected neural network, can be replaced by other networks, e.g., LSTM.
We can also drop the predictor in applications where only the reconstruction is needed, as we will introduce in experiment 1.

\subsection*{\textbf{Buffers}}
Every neural network that needs to be updated during its application owns a limited novelty buffer and an unlimited familiarity buffer. 
When true target values are provided, the MSE can be used as a criterion to be compared to the preset threshold.
The samples with a small MSE are stored in the familiarity buffer because the trained network has learned to cope with them before. 
The samples with a large MSE are stored in the novelty buffer.
The network needs to be retrained based on these novelties to learn new knowledge.
After updating, a validation error can be calculated using the familiarities to estimate whether the network can still retain the old knowledge acquired previously.
How to deal with a poor update result remains an open question and needs to be further discussed.
We empty both buffers after finishing an update.

\subsection*{\textbf{Threshold}}
Each model that needs to be updated owns a threshold.
As shown in Fig.~\ref{fig:details}, \textbf{Threshold\_a} is for the autoencoder and \textbf{Threshold\_p} is for the predictor.
The value of the threshold determines how new samples are classified.
The smaller the threshold, the more samples are likely to be labeled as novelties, where more well-learned knowledge will be re-learned, thus leading to unnecessary updates.
A larger threshold could cause inefficient updates because too many novelties are misclassified.
We suggest that the threshold value should be adjusted dynamically depending on the training results.
\begin{figure}[t]
	\centering
	\includegraphics[width=0.45\textwidth]{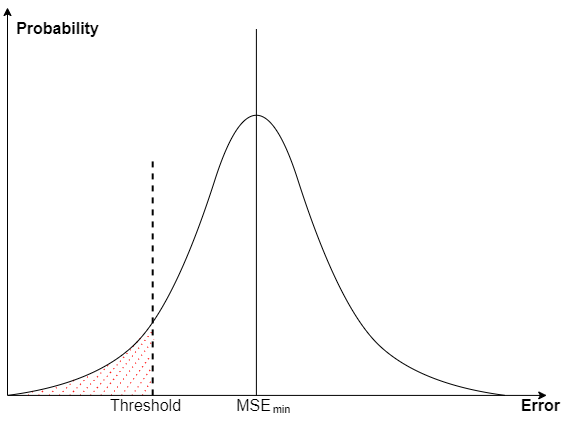}
	\caption{A probability distribution of errors for the data in the novelty buffer (or training dataset) and the familiarity buffer (or test dataset). $\mathit{MSE}_{\min}$ indicates the minimum MSE after learning.}
	\label{fig:threshold}
\end{figure} 

Fig.~\ref{fig:threshold} illustrates a distribution of errors for all samples after learning.
$\mathit{MSE}_{\min}$ refers to the minimum MSE obtained by minimizing a loss function, which can also be viewed as the mean of the distribution.
The distribution has a lower mean and a lower variance indicating the model learns better on the given dataset.
In this article, we adjust the threshold by
\begin{eqnarray}
\centering
\small
\begin{aligned}
\mathit{Threshold} = \alpha \times \mathit{MSE}_{\min}, \text{with}\ \alpha > 0,
\end{aligned}
\label{formula:threshold}
\end{eqnarray}
where $\alpha$ is a fixed threshold factor and the $\mathit{MSE}_{\min}$ is re-estimated after each update. 

\subsection*{\textbf{Update}}
The updating method and the trigger condition play a crucial role in the CLeaR framework. 
As mentioned in the Section about related work, retraining on old tasks should be constrained due to reasons, such as privacy protection or data storage overhead. 
Therefore, we adopt Online-EWC, which penalizes the loss when the overlapped significant weights are changed while learning new tasks. 
We adapt the notation given in~\cite{huszar2017quadratic} to explain EWC and Online-EWC in the context of the CLeaR framework.

\subsubsection*{EWC}
The goal of EWC is to approximate Bayesian posteriors over model parameters given tasks.
In CLeaR, data is always split according to two kinds of tasks, the known task ($T_k$) and the unknown task ($T_u$).
$T_k$ refers to what the neural network has already learned. 
The corresponding dataset $D_{k, T-1}$ is a combination of the datasets that were stored in the novelty and the familiarity buffers during the $T-1$th update. 
Note that novelties in $D_{k, T-1}$ have already been learned before the $T$th update. 
Therefore, it belongs to the known task.
Comparably, $T_u$ refers to what the network will learn at the $T$th update.
The corresponding $D_{u, T}$ refers only to the dataset stored in the novelty buffer for triggering the $T$th update. 
Chronologically, the $T-1$th update occurs before the $T$th update.
From Eq.~\ref{formula:prior_focused}, we can get the posterior after learning $T_u$, as:
\begin{eqnarray}
\centering
\small
\begin{aligned}
P\left (\Theta|D_{u, T}, D_{k, T-1}\right) = \frac{P\left(D_{u, T}|\Theta\right)P\left(\Theta|D_{k, T-1}\right)}{P\left(D_{u, T}|D_{k, T-1}\right)},
\end{aligned}
\label{formula:posterior}
\end{eqnarray}
By converting Eq.~\ref{formula:posterior} into a logarithm, we will get
\begin{eqnarray}
\centering
\small
\begin{aligned}
\log P\left (\Theta|D_{u, T}, D_{k, T-1}\right) &= \log \left(D_{u, T}|\Theta\right) \\&+ \log P\left(\Theta|D_{k, T-1}\right) \\&- \log P\left(D_{u, T}|D_{k, T-1}\right).
\end{aligned}
\label{formula:log_posterior}
\end{eqnarray}
The term $\log P(D_{u,T}|\Theta)$ is generally tractable through minimizing an MSE loss with respect to $\Theta$ and the dataset $D_{u,T}$.
In the case where the trained network can perform very well on the known task $T_k$, we have gotten the optimal parameters $\Theta^{*}_{k,T-1}$, which makes the gradient of $-\log P(\Theta|D_{k,T-1})$ with respect to $\Theta$ equal to 0. 
Therefore, the $-\log P(\Theta|D_{k,T-1})$ can be estimated using 2nd order Taylor series around $\Theta^{*}_{k,T-1}$:
\begin{eqnarray}
\centering
\small
\begin{aligned}
-\log P\left(\Theta|D_{k, T-1}\right) &\approx \frac{1}{2}\Delta\left(\Theta\right)^\mathrm{T} H\left(\Theta^{*}_{k, T-1}\right) \Delta\left(\Theta\right) \\&+ \mathrm{constant},
\end{aligned}
\label{formula:taylor}
\end{eqnarray}
where $\Delta\left(\Theta\right)=\Theta - \Theta^{*}_{k, T-1}$.
$H(\Theta^{*}_{k, T-1})$ is the Hessian of $-\log P(\Theta|D_{k, T-1})$ with respect to $\Theta$, evaluated at the optimum $\Theta^{*}_{k,T-1}$. 
Furthermore, we can approximate the Hessian as:
\begin{eqnarray}
\centering
\small
\begin{aligned}
H\left(\Theta^{*}_{k,T-1}\right) \approx N_k \cdot \mathbf{F}\left(\Theta^{*}_{k,T-1}\right) + H_{prior}\left(\Theta^{*}_{k,T-1}\right),
\end{aligned}
\label{formula:hessian}
\end{eqnarray}
where $N_k$ is the number of samples in $D_{k,T-1}$, and $\mathbf{F}(\Theta^{*}_{k,T-1})$ is the empirical Fisher information matrix on the known dataset $D_{k,T-1}$, and $H_{prior}(\Theta^{*}_{k,T-1})$ is the Hessian of the negative log prior with respect to $\Theta$.
EWC estimates the Fisher information matrix in a high-dimensional space as a diagonal matrix, i.e., the non-diagonal elements are zero~\cite{kirkpatrick2017overcoming}. 
The diagonal Fisher information values are denoted by $F_k^i$ concerning the $i$th parameter in the network.
Thus the formula of EWC is
\begin{eqnarray}
\centering
\small
\begin{aligned}
\Theta^{*}_{k, T} =& \argmin_{\Theta} \Bigl \{ -\log P\left(D_{u,T}|\Theta\right) \\+& \frac{1}{2}\sum_{i} \left(\sum_{t=1}^{T-1}\lambda_t F_{k,t}^{i} + \lambda_{prior}\right)\left(\theta^i - \theta_{k, T-1}^{*,i}\right)^2 \Bigl \},
\end{aligned}
\label{formula:ewc}
\end{eqnarray}
where the update has been done $T-1$ times already. 
$D_{u,T}$ is the dataset stored in the novelty buffer for the unknown task at the $T$th update. 
$F_{k,t}^i$ is the diagonal Fisher information with respect to the parameter $i$, estimated by the dataset $D_{k,t}$ stored in both buffers after the $T$th update. 
$\theta_{k, T-1}^{*,i}$ is the optimal parameter $i$ learned from the dataset $D_{k, T-1}$ after the $T-1$th update.
$\lambda_t$ and $\lambda_{prior}$ are hyperparameters for EWC.
Note that all of the diagonal Fisher information $F_{k,t}$ in Eq.~\ref{formula:ewc}, obtained after every update, must be stored. 

\subsubsection*{Online-EWC}
The regularization term in EWC can be replaced by one Gaussian approximation to the whole posterior of all previous tasks, as proven in~\cite{huszar2017quadratic}. 
Based on this derivation, Online-EWC is proposed in~\cite{schwarz2018progress}, as:
\begin{eqnarray}
\centering
\small
\begin{aligned}
\Theta^{*}_{k, T} =& \argmin_{\Theta} \Bigl \{ -\log P\left(D_{u,T}|\Theta\right) \\+&  \frac{1}{2}\sum_{i} \lambda \widetilde{F}_{k,T-1}^{i}(\theta^i - \theta_{k,T-1}^{*,i})^2 \Bigl \},
\end{aligned}
\label{formula:o_ewc}
\end{eqnarray}
where $\widetilde{F}_{k,T}^{i} = \gamma\widetilde{F}_{k,T-1}^{i} + F_{k,T}^{i}$ and $\widetilde{F}_{k,1}^{i}=F_{k,1}^{i}$. 
$\gamma$ is a hyperparameter governing the contribution of previous tasks and not larger than 1. 
Therefore, $\widetilde{F}_{k,T-1}^{i}$ contains diagonal Fisher information of all updates which precede the $T$th update. 

By contrast with EWC, Online-EWC requires less space to store diagonal Fisher information and optima of each previous task. 
Another similar algorithm is EWC++ introduced in~\cite{chaudhry2018riemannian}, where the diagonal Fisher information is defined as: $\widetilde{F}_{k,T-1}^{i}=\alpha\widetilde{F}_{k, T-1}^{i} + (1-\alpha)F_{k,T}^{i}$ with $0<\alpha<1$. 
In our experiments, Online-EWC is selected for updates within the CLeaR framework.

\subsubsection*{Trigger condition}
The trigger condition for updating the model is another customizable parameter. 
In our realization of the framework, the update will be triggered when the finite novelty buffer is filled. 
The smaller the size of this novelty buffer, the more easily this buffer gets full. 
On the one hand, fewer data in a full buffer is available for updating, which might lead to a failed update.
On the other hand, more updates will result in increasing computational overhead.
Moreover, note that the hyperparameter $\gamma$ of Online-EWC is not larger than 1, which causes a gradual decay of the previous Fisher information when the number of updates increases.
If the size of the buffer is too large, updating the network will be delayed until the buffer is full.
The delay also results in the fact that the threshold can not be adjusted frequently.

\section*{Experiments}
In this section, two sets of experiments will be performed and analyzed. 
An unlabeled artificial dataset is generated for experiment 1, where we analyze the effects of two hyperparameters of CLeaR, the novelty buffer size and the threshold factor.
We implement experiment 2 on a labeled dataset to predict wind power generation measured in 10 European wind farms~\cite{gensler2016} using supervised learning. 
In this experiment, we assess the CLeaR framework's performance in wind power prediction applications. Both experiments will adopt three evaluation metrics: fitting error, prediction accuracy, and forgetting ratio. 
The remainder of the section will introduce the datasets, the experimental setup, the evaluation criteria, and analyze the results.

\subsection*{\textbf{Dataset}}
\paragraph*{Artificial dataset} is a seven-dimensional periodic unsupervised dataset, i.e., $D=\left [ \mathbf{x}_1,\dots,\mathbf{x}_7 \right ]$. 
The underlying generation model is shown as follows:
\begin{eqnarray}
\centering
\small
\begin{aligned}
x_{n}(t) &\sim{} N(mean_{n}(t), var_{n}(t)), \\
mean_{n}(t) &= A_{m} \times \left | \sin(\frac{\pi \times (t+p_{n})}{\mathrm{T}})\right |, \\
var_{n}(t) &= A_{v} \times \left |\sin(\frac{\pi \times (t+p_{n})}{\mathrm{T}})\right |, \\
p_{n} &\sim{} N(0, 1),
\end{aligned}
\label{formula:artificial_ts}
\end{eqnarray}
where $x_n(t)$ is sampled from a Gaussian distribution with a time-dependent mean and variance and denotes the $t$th data of the sequence $\mathbf{x}_n$.
The mean and variance are sampled from an absolute value of a sinusoidal function at the $t$th point in time with a period $\mathrm{T}$ and a phase $p_n$. 
The phase $p_n$ is randomly generated with a standard normal distribution. 
Hyperparameters $A_{m}$ and $A_{v}$ refer to the amplitude of the sinusoidal function. 

In order to generate a dataset containing daily and yearly periodical changes, we adopt two generation models, $x_n^{d}(t)$ and $x_n^{y}(t)$, with two periods $\mathrm{T}^d$ and $\mathrm{T}^y$ respectively.
The period $\mathrm{T}^d$ is $24$ to simulate the fluctuation within one day ($24$ hours), and the period $\mathrm{T}^y$ of $x_n^{y}(t)$ is $8760$ to mimic the periodic changes over a year ($8760$ hours).
The resulting time series data is a sum of the two models, i.e., $x_n(t) = x_n^{d}(t)+x_n^{y}(t)$.

\begin{figure}[t!]
	\centering
	\subfigure[$\mathbf{x}_1$ over 48 hours (2 days)]{
		\includegraphics[width=.442\textwidth]{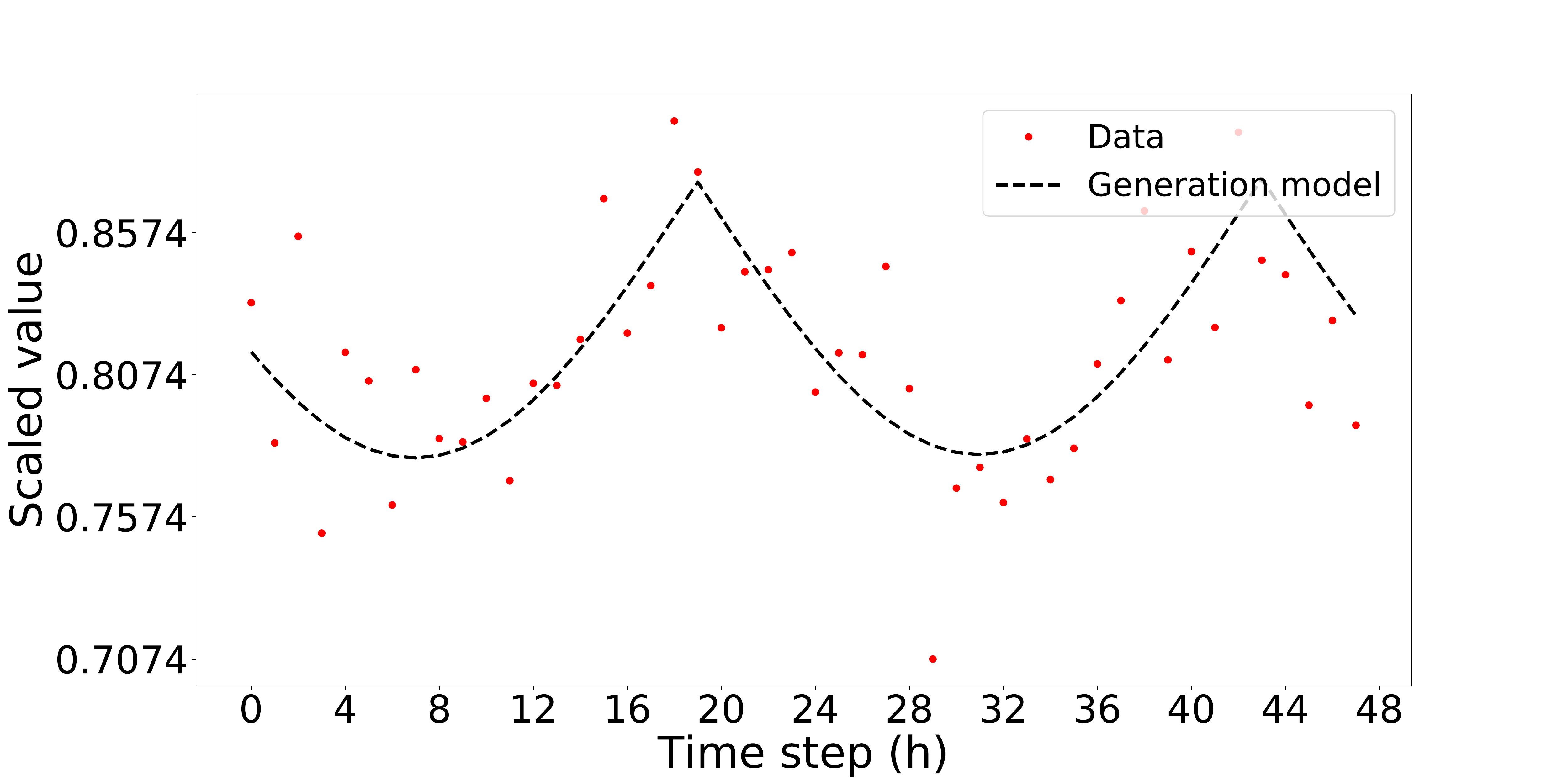}
		\label{subfig:ts_2d}
	}\\
	\subfigure[$\mathbf{x}_1$ over 720 hours (30 days)]{
		\includegraphics[width=.442\textwidth]{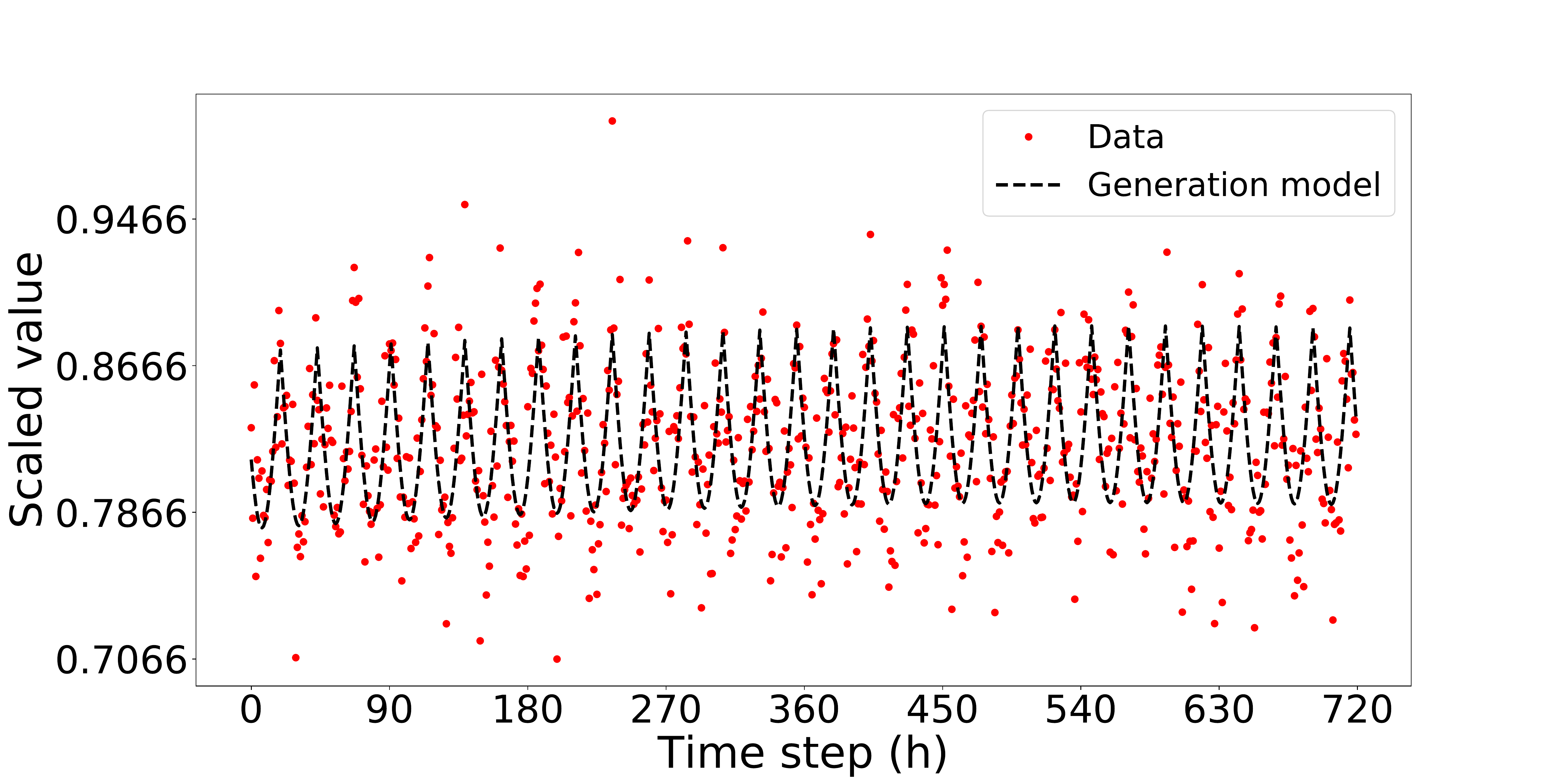}
		\label{subfig:ts_1m}
	}
	\\
	\subfigure[$\mathbf{x}_1$ over 17520 hours (2 years)]{
		\includegraphics[width=.442\textwidth]{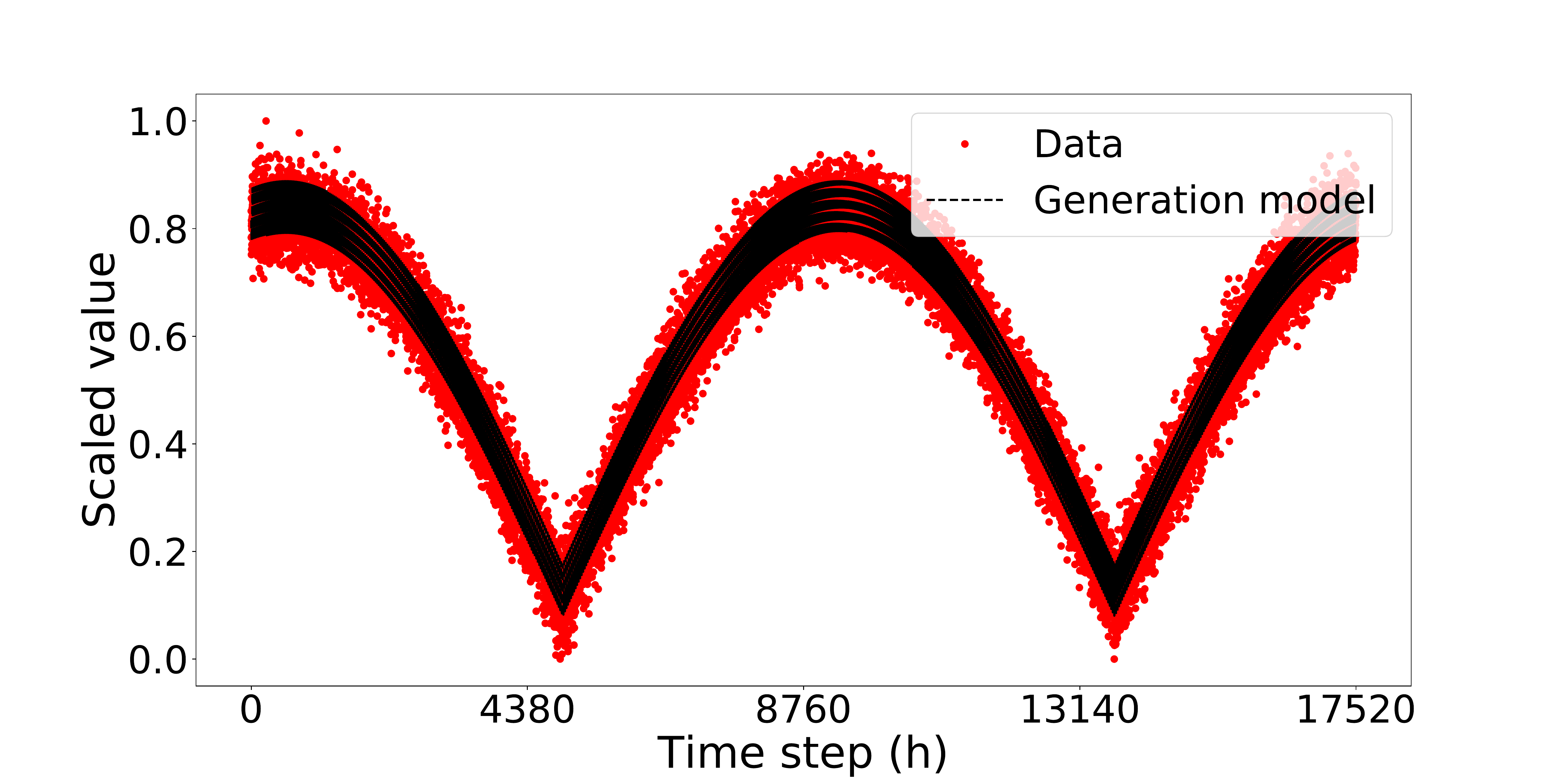}
		\label{subfig:ts_2y}
	}
	\caption{The three subfigures illustrate the fluctuation of the first dimension sequence $\mathbf{x}_1$ over 48 hours (2 days), 720 hours (30 days), and 17520 hours (2 years), respectively. The underlying generation model is marked in black, and generated data is in red. The generation model of $\mathbf{x}_1$ is a sum of two sub-models $\mathbf{x}_1^{d}$ and $\mathbf{x}_1^{y}$, which have different periods $\mathrm{T}^d$ (24 hours) and $\mathrm{T}^y$ (8760 hours) respectively.}
	\label{fig:artifical_ts}
\end{figure}

Fig.~\ref{fig:artifical_ts} displays the sequence $\mathbf{x}_1$ (marked in red) and the corresponding generating process (marked in black) in three different time slots.
Note that the first 48 samples of Fig.~\ref{subfig:ts_1m} are displayed in Fig.~\ref{subfig:ts_2d}.
Similarly, the first 720 samples of Fig.~\ref{subfig:ts_2y} are also shown in Fig.~\ref{subfig:ts_1m}.
The periodical statistic fluctuation displayed in Fig.~\ref{fig:artifical_ts} can simulate features of non-stationary data, such as temperature, which usually peaks at noon and falls at night.

\paragraph*{Wind power dataset}\cite{gensler2016} contains seven meteorological features and hourly averaged wind power generation data measured from European wind farms for two years in a row. 
The features are 24-hour-ahead meteorological forecasts using the European Centre for Medium-Range Weather Forecasts model~\cite{ecmwfweb}, including (1) wind speed in 100m height, (2) wind speed in 10m height, (3) wind direction (zonal) in 100m height, (4) wind direction (meridional) in 100m height, (5) air pressure, (6) air temperature, and (7) humidity.

The power generation time series are normalized by the respective rated capacity of the wind farm for easy, scale-free comparison.
All features are normalized to the range between 0 and 1.
When no power has been generated longer than 24 hours, time points are removed in the pre-processing phase. 
The dropped time points can be viewed as exceptions that need to be processed manually in real-world applications. 

Both datasets are available on our department website~\cite{gensler2016} or by contacting the corresponding author. 
Researchers can re-implement our experiments with the datasets.

\subsection*{\textbf{Experimental setup}}
The artificial dataset and the wind power dataset are split into three subsets according to three phases: \textbf{a warm-up phase}, \textbf{an update phase}, and \textbf{an evaluation phase}. 
In the warm-up phase, the model is pre-trained on the first 1000 samples, containing only partial information that can describe the current task but not future tasks.
The model monitors a data stream with the following 10000 samples in the update phase, simulating a theoretically infinite data stream in a real application scenario. 
Here, we assume that 10000 samples can provide enough information to describe the real distribution of samples because 10000 samples span 10000 hours, covering a complete period.
The model will be retrained based on the novelty buffer and validated on the familiarity buffer once updating is triggered. 
The updated model is finally evaluated in the evaluation phase with the following 1000 samples.

We implement the following three CLeaR instances and a baseline model to compare their performances in experiments:
\begin{itemize}
	\item Instance$_{\textbf{A}}$: Its models are trained on the warm-up dataset in supervised learning mode and then directly evaluated in the evaluation phase without updating. 
	The experimental results can reflect the disadvantages due to the lack of samples. 
	Instance$_{\textbf{A}}$ is viewed as a \textbf{lower bound} of models' performance.
	\item Instance$_{\textbf{B}}$: Its models are pre-trained in the warm-up phase, as Instance$_{\textbf{A}}$. 
	Then the models will be updated in the update phase by using \textbf{fine-tuning}.
	Fine-tuning allows the pre-trained models to learn a new dataset, which they were not originally trained on, by slightly adjusting all unfrozen parameters. 
	The re-training process is monitored by early stopping with 30 epochs of patience to avoid overfitting. 
	Early stopping will stop the re-training process if the loss is no longer decreasing in 30 epochs.
	The optimal models are the ones with the lowest loss before stopping.
	Identically, the updated Instance$_{\textbf{B}}$ will be evaluated in the evaluation phase after the update phase.
	\item Instance$_\textbf{C}$: It is similar to Instance$_{\textbf{B}}$, but uses \textbf{Online-EWC} without being monitored by early stopping for updating. 
	It will be evaluated in the evaluation phase as other models.
	\item Baseline model: It is a common deep neural network model with the same structure as the neural network models in the above CLeaR instances. The baseline model is traditionally trained with 11000 samples (i.e., the samples of the warm-up phase and the update phase). Dropout layers with a dropout rate of 0.2 are used during training to avoid overfitting. 
	It will also be evaluated in the evaluation phase.
\end{itemize}

Two sets of experiments are conducted on the artificial dataset and the wind power dataset, respectively. 
\begin{itemize}
	\item Experiment \textbf{1} is based on the artificial dataset to analyze the correlation between the CLeaR framework's performance and two framework-related hyperparameters, i.e., the novelty buffer size and the threshold factor. 
	This unsupervised experiment aims to extract latent representations and reconstruct the inputs.
	The Instance$_{\textbf{C}}$ (only an autoencoder involved) and the baseline model (an autoencoder) are implemented. 
	
	The grid search range of the framework's hyperparameters is shown in Table~\ref{tab:exp_1_hyperparamters}, including 56 available pairs of parameter values. 
	The experiment with each pair of parameters is repeated 20 times with different initialization of the autoencoder.
	Identically, the baseline model is also repeated 20 times.
	The architecture of the autoencoder and the training setting for experiment 1 are shown in Tables~\ref{tab:autoencoder_exp_1} and~\ref{tab:trainingsetting_exp_1}.
	\begin{table}[t!]
    	\caption{The grid search range of framework-related parameters for experiment 1. }
    	\begin{tabular}{c@{\hskip 0.1mm}c@{\hskip 0.2mm}}
    		\hline
    		Hyperparameter &  Value\\ \hline
    		Novelty buffer & $\left\{400, 600, 800, 1000, 1200, 1400, 1600, 1800\right\}$ \\
    		Threshold factor & $\left\{0.5, 0.65, 0.8, 0.95, 1.1, 1.25, 1.4 \right\}$ \\ \hline
    	\end{tabular}
    	\label{tab:exp_1_hyperparamters}
    \end{table}
    
    \begin{table}[t!]
    	\caption{The architecture of the autoencoder in experiments 1. Dropout layers are used with a dropout rate of 0.2. The slope parameter of the LeakyReLu is set to 0.05.}
    	\begin{tabular}{cccc}
    		\hline
    		Layer & Size & Activation  \\ \hline
    		Encoder-Input & 7 & None \\
    		Encoder-1 & 7$\times$32 & LeakyReLu\\
    		Encoder-2 & 32$\times$16 & LeakyReLu\\
    		Encoder-3 & 16$\times$8 & LeakyReLu\\
    		Encoder-4 & 8 $\times$4 & LeakyReLu\\
    		Decoder-1 & 4 $\times$8 & LeakyReLu\\
    		Decoder-2 & 8$\times$16 & LeakyReLu\\
    		Decoder-3 & 16$\times$32 & LeakyReLu\\
    		Decoder-4 & 32$\times$7 & LeakyReLu\\
    		Decoder-Output & 7 &None\\ \hline
    	\end{tabular}
    	\label{tab:autoencoder_exp_1}
    \end{table} 

    \begin{table}[t!]
    	\caption{Training setting for experiment 1. Epochs denote the number of iterations for training the autoencoder. The batch size is the size of a mini-batch. Phase 1 and phase 2 indicate the warm-up phase and the update phase respectively. The experiments of training baseline model adopt the setting of the phase 1.}
    	\begin{tabular}{cccc}
    		\hline
    		Hyperparameter & Value \\ \hline
    		Optimizer & Adam \\
    		Epochs in phase $1$ & 512 \\
    		Epochs in phase $2$ & 512 \\
    		Batch size in phase $1$ & 32 \\
    		Batch size in phase $2$ & 16 \\ 
    		Online EWC $\gamma$ & 0.9 \\
    		Online EWC $\lambda$ & 200\\ \hline
    	\end{tabular}
    	\label{tab:trainingsetting_exp_1}
    \end{table}
    
	\item Experiment \textbf{2} aims to evaluate all three CLeaR instances and the baseline model in the real-world application of wind power generation forecast based on 10 European wind farms datasets. 
	The adopted model consists of an autoencoder and a deep neural network, as proposed in Fig.~\ref{fig:details}.
	Its architecture parameters are empirically selected and kept identical for a fair comparison, referring to Table~\ref{tab:structure_exp_2}.
	The parameters for the framework and Online EWC algorithm are selected using grid search from the range in Table~\ref{tab:hyperparamters_exp_2}.
	The training setting for experiment 2 is shown in Table~\ref{tab:trainingsetting_exp_2}.
	\begin{table}[t!]
    	\caption{Architecture parameters of the model for experiment 2. Dropout layers are used with a dropout rate selected by means of grid search. Latent size refers to the bottleneck dimension in the autoencoder. The slope parameter of the LeakyReLu is set to 0.05.}
    	\begin{tabular}{cccc}
    		\hline
    		Layer & Size & Activation  \\ \hline
    		Encoder-Input & 7 & None \\
    		Encoder-1 & 7$\times$32 & LeakyReLu\\
    		Encoder-2 & 32$\times$16 & LeakyReLu\\
    		Encoder-3 & 16$\times$8 & LeakyReLu\\
    		Encoder-4 & 8 $\times$Latent size & LeakyReLu\\
    		Decoder-1 & Latent size $\times$8 & LeakyReLu\\
    		Decoder-2 & 8$\times$16 & LeakyReLu\\
    		Decoder-3 & 16$\times$32 & LeakyReLu\\
    		Decoder-4 & 32$\times$7 & LeakyReLu\\
    		Decoder-Output & 7 &None\\
    		DNN-1 & Latent size$\times$96 &Tanh\\
    		DNN-2 & 96$\times$64 &Tanh\\
    		DNN-3 & 64$\times$32 &Tanh\\
    		DNN-4 & 32$\times$16 &Tanh\\
    		DNN-5 & 16$\times$8 &Tanh\\
    		DNN-Output & 8$\times$1 &None\\ \hline
    	\end{tabular}
    	\label{tab:structure_exp_2}
    \end{table} 
    \begin{table}[t!]
    	\caption{The grid search range of hyperparameters for experiment 2. The parameter $\lambda_a$ is the lambda value for the autoencoder, and $\lambda_p$ is for the predictor. Latent size is the dimension of the bottleneck in the autoencoder.}
    	\begin{tabular}{cccc}
    		\hline
    		Hyperparameter & Value \\ \hline
    		Dropout & $\left \{0, 0.005 \right \} $\\
    		Novelty buffer size & $\left \{512, 1024 \right \}$ \\
    		Latent size & $\left \{4, 5\right \}$\\
    		Threshold factor & $\left\{ 0.75\right\}$\\
    		Online EWC $\gamma$ & $\left \{0.7, 0.8, 0.9\right \}$\\
    		Online EWC $\lambda_a$ & $\left \{200\right \}$\\
    		Online EWC $\lambda_p$ & $\left \{600, 800\right \}$\\ \hline
    	\end{tabular}
    	\label{tab:hyperparamters_exp_2}
    \end{table}
	\begin{table}[t!]
    	\caption{Training setting for the experiment 2. The parameter epoch$_a$ denotes the number of iterations for training the autoencoder and epoch$_p$ for training the predictor. The batch size is the size of a mini-batch. Phase 1 and phase 2 indicate the warm-up phase and the update phase respectively.}
    	\begin{tabular}{cccc}
    		\hline
    		Hyperparameter & Reference figure \\ \hline
    		Optimizer & Adam \\
    		Epochs$_a$ in phase $1$ & 256 \\
    		Epochs$_p$ in phase $1$ & 256 \\
    		Batch size in phase $1$ & 32 \\
    		Epochs$_a$ in phase $2$ & 128 \\
    		Epochs$_p$ in phase $2$ & 256 \\
    		Batch size in phase $2$ & 16 \\ \hline
    	\end{tabular}
    	\label{tab:trainingsetting_exp_2}
    \end{table}
\end{itemize}

\subsection*{\textbf{Metrics}}
Models are evaluated in terms of fitting error, prediction error, and forgetting ratio. 
\paragraph*{\textbf{Fitting error}} indicates how well the instance fits all seen samples after the update phase.
Updating a CLeaR instance by mini-batch data might lead to a local minimum during the updating process. 
Eventually, the instance fits only a specific subset rather than all seen data. 
Such effects are measured by calculating the MSE on the 11000 samples, 1000 of which are from the warm-up phase, and the rest 10000 are from the update phase.
Therefore, a lower fitting error reflects that more knowledge is finally accumulated.
\paragraph*{\textbf{Prediction error}} reflects the ability of CLeaR instances to perform predictions on previously unseen data.
It is calculated with 1000 samples in the evaluation phase. 
The overfitting problem often exists due to neural networks' powerful learning ability, where a test error is much larger than a training error. 
Regularization techniques, such as dropout and early stopping, are used to avoid overfitting during training.
The prediction error can tell us whether the instance has already fallen into a local optimum after several updates.
Also, the comparison between Instance$_{\textbf{A}}$ and Instance$_{\textbf{C}}$ in experiment 2 shows how important accumulating data and updating the models are for improving prediction accuracy, especially under the condition of a limited pre-training dataset.
\paragraph*{\textbf{Forgetting ratio}} measures how much old knowledge a model forgets after learning new tasks. 
In~\cite{he2020continuous}, He et al. compared the forgetting ratio to average test error and demonstrated that the forgetting ratio could reflect the severity of the forgetting problem. 
The formula is
\begin{eqnarray}
\centering
\small
\begin{aligned}
\mathit{forgetting\ ratio} = \frac{\max(0, \mathit{L_{warm\_up}}^{2} - \mathit{L_{warm\_up}}^{1})}{\mathit{L_{warm\_up}}^{1}},
\end{aligned}
\label{formula:forgetting_ratio}
\end{eqnarray}
where \textit{$L_{warm\_up}^{1}$} indicates the MSE on the warm-up dataset at the end of the warm-up phase and \textit{$L_{warm\_up}^{2}$} indicates the error on the same dataset at the end of the update phase, and $\max(x_1, x_2)$ returns the larger one of either $x_1$ or $x_2$. 
The increment of the error for the same task describes the model's forgetfulness after learning new tasks. 
The comparison is performed only between the Instance$_{\textbf{B}}$ and the Instance$_{\textbf{C}}$ in experiment 2 because the Instance$_{\textbf{A}}$ and the baseline model are not updated.


\subsection*{\textbf{Results of the experiment 1}}
\begin{figure*}[t!]
	\centering
	\includegraphics[width=0.89\textwidth]{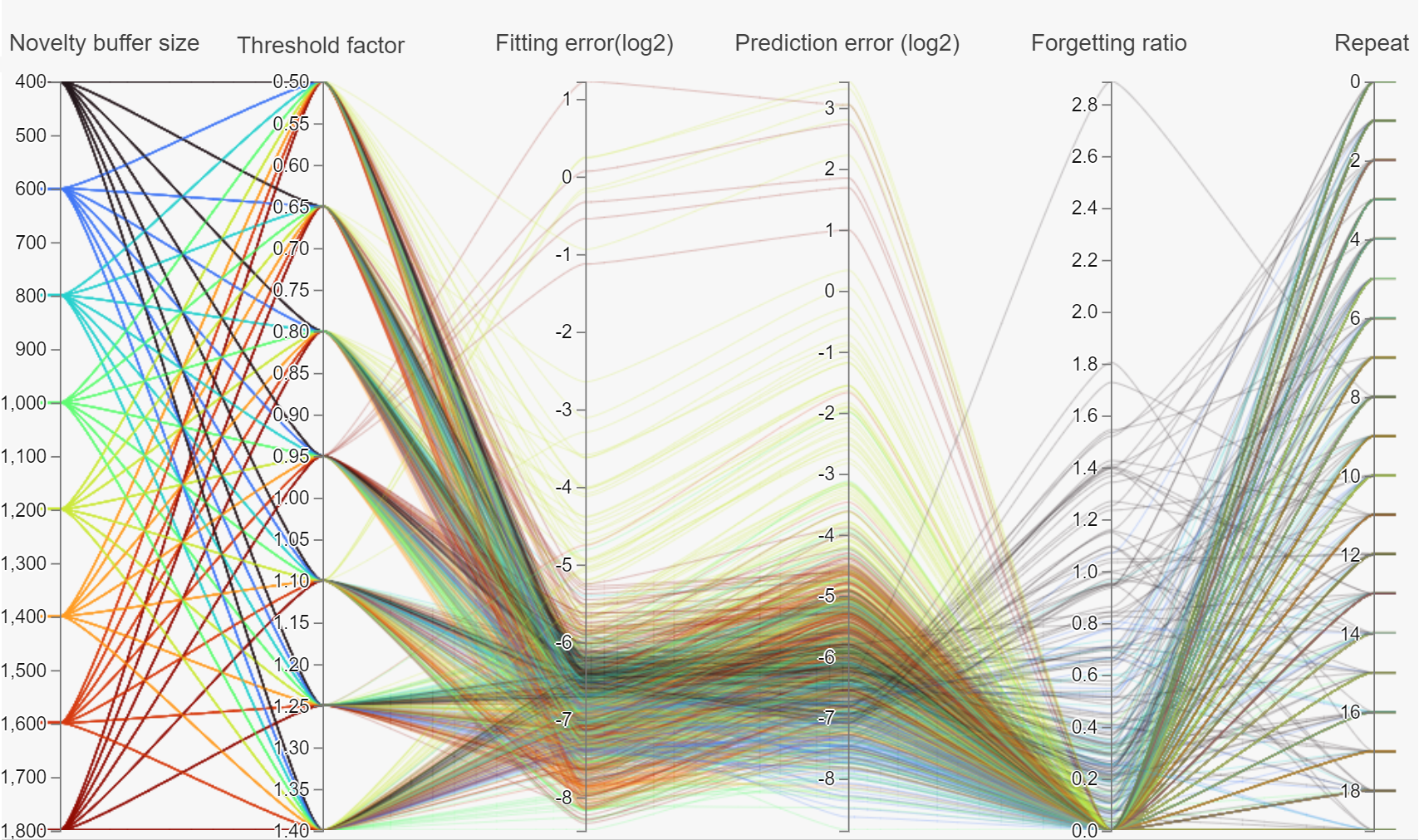}
	\caption{The parallel coordinates plot illustrates correlations between the frameworks' hyperparameters (novelty buffer size and threshold factor) and the evaluation metrics (fitting error, prediction error, and forgetting ratio) based on 1120 results obtained in experiment 1. 56 pairs of hyperparameters are repeated 20 times. The best fitting error and the best prediction error of the baseline model in 20 repeated experiments are -4.82 and -6.58.}
	\label{fig:parallel_coordinates_plot}
\end{figure*}

Fig.~\ref{fig:parallel_coordinates_plot} illustrates the correlation between the frameworks' hyperparameters (novelty buffer size and threshold factor) and the evaluation metrics (fitting error, prediction error, and forgetting ratio) based on 1120 results obtained in experiment 1.
\textbf{The lower the evaluation metric is, the better the model performs}.
Note that the binary logarithm of the fitting error and prediction error are plotted in Fig.~\ref{fig:parallel_coordinates_plot}.
Moreover, the binary logarithm of the best fitting error and the best prediction error of the baseline model are -4.82 and -6.58, respectively.

Regarding the metric of fitting error, 1089 of the 1120 CLeaR models (97.2\%) perform better than the best baseline model.
This result indicates that the CLeaR framework can accumulate knowledge continually and effectively.
Besides, it shows that the CLeaR framework's continual fitting ability is relatively robust to both framework-related hyperparameters.

Regarding the prediction error, 360 of the 1120 CLeaR models (32.1\%) outperform the best baseline model.
253 of the 360 models have a threshold factor greater than or equal to 0.95. 
159 of the 253 models have a novelty buffer size lower than or equal to 1000. 
Furthermore, 455 CLeaR models (40.8\%) perform worse than the worst baseline model, whose prediction error is -5.84.
267 of the 455 models have a threshold factor lower than or equal to 0.95. 
203 of the 267 models have a novelty buffer size greater than or equal to 1200. 
On the one hand, the experimental results are in accordance with our expectation, i.e., continual accumulation of meaningful knowledge can improve neural networks' prediction abilities.
To a certain extent, the CLeaR framework can even predict non-stationary data more accurately than a neural network trained with sufficient historical samples.
On the other hand, our findings indicate that the CLeaR framework's prediction ability is susceptible to its hyperparameter values.
In the update phase, the average update frequency of the 360 models that outperform the best baseline model is 10.2 times, while the average update frequency of the 455 models that perform worse than the worst baseline model is 6.87 times.
A smaller novelty buffer gets filled more easily so that updating is triggered more frequently.
Besides, a higher threshold can result in that only high-entropy data is stored in the novelty buffer.
We conclude that timely learning of high-entropy data can effectively improve the prediction accuracy of neural networks for non-stationary data.

We only analyze the forgetting ratio of these CLeaR models because the forgetting problem does not happen with the baseline model.
174 of the 1120 CLeaR models (15.5\%) obtain a forgetting ratio greater than 0.1.
172 of the 174 models have a novelty buffer size lower than or equal to 1000.
Combined with the conclusion concerning the prediction error, we find that a smaller novelty buffer can trigger updating more often, which can decrease the prediction error but also increase the forgetting ratio.
The main reason can be that the hyperparameter $\gamma$ of Online EWC is set to 0.9 (see Table~\ref{tab:trainingsetting_exp_1}), which leads to a decay of the Fisher information regarding the previous tasks after each update.
Therefore, how to adjust the hyperparameters and how to supervise the updating process will be one of the key points in our further research. 
Otherwise, it will always be faced with a trade-off between prediction error and forgetting ratio.

Fig.~\ref{fig:parallel_coordinates_plot} is plotted using HiPlot~\cite{hiplot}.
The raw data of the results is available by contacting the corresponding author.
\subsection*{\textbf{Results of the experiment 2}}
\begin{table}[t!]
	\caption{The fitting errors of the \textbf{autoencoders} in the three CLeaR instances and the baseline model. $ae$ refers to the autoencoder.}
	\begin{tabular}{c@{\hskip 2.5mm}c@{\hskip 2.5mm}c@{\hskip 2.5mm}c@{\hskip 2.5mm}|c@{\hskip 2.5mm}}
		\hline
		Dataset&\multicolumn{4}{c}{Fitting error (e-2)}\\ 
		\cline{2-5}
		&Instance$_{\textbf{A}_{ae}}$ & Instance$_{\textbf{B}_{ae}}$ & Instance$_{\textbf{C}_{ae}}$ & Baseline$_{ae}$\\
		\hline
		WF 1 & 2.098 & $\mathbf{0.369}$ & 0.502 & 1.505 \\
		\hline
		WF 2 & 2.068 & $\mathbf{0.348}$ & 0.440 & 1.579 \\
		\hline
		WF 3 & 2.292 & $\mathbf{0.423}$ & 0.513 & 1.634 \\
		\hline
		WF 4 & 2.669 & 0.988 & $\mathbf{0.863}$ & 1.449 \\
		\hline
		WF 5 & 1.421 & 0.516 & $\mathbf{0.471}$ & 1.549 \\
		\hline
		WF 6 & 2.456 & 1.054 & $\mathbf{0.761}$ & 1.065 \\
		\hline
		WF 7 & 2.129 & $\mathbf{0.647}$ & 0.931 & 1.441 \\
		\hline
		WF 8 & 2.149 & $\mathbf{0.434}$ & 0.662 & 1.448 \\
		\hline
		WF 9 & 2.370 & $\mathbf{0.773}$ & 0.793 & 1.340 \\
		\hline
		WF 10 & 1.996 & 0.678 & $\mathbf{0.411}$ & 1.406 \\	
		\hline
		Mean & 2.165 & $\mathbf{0.623}$ & 0.635 & 1.442 \\
		\hline		
	\end{tabular}
	\label{tab:fitting_ae}
\end{table}
\begin{table}[t!]
	\caption{The fitting errors of the \textbf{predictor} in the three CLeaR instances and the baseline model.}
	\begin{tabular}{c@{\hskip 3.5mm}c@{\hskip 3.5mm}c@{\hskip 3.5mm}c@{\hskip 3.5mm}|c@{\hskip 3.5mm}}
		\hline
		Dataset&\multicolumn{3}{c}{Fitting error (e-2)}\\
		\cline{2-5}
		&Instance$_{\textbf{A}}$ & Instance$_{\textbf{B}}$ & Instance$_{\textbf{C}}$ & Baseline\\
		\hline
		WF 1 & 4.864 & 3.468 & $\mathbf{2.304}$ & 2.574 \\
		\hline
		WF 2 & 2.695 & 2.120 & $\mathbf{1.167}$ & 1.473 \\
		\hline
		WF 3 & 3.517 & 9.172 & $\mathbf{1.769}$ & 2.840 \\
		\hline
		WF 4 & 4.867 & 5.457 & $\mathbf{3.682}$ & 3.327 \\
		\hline
		WF 5 & 10.347& 12.989 & $\mathbf{6.931}$ & 6.835 \\
		\hline
		WF 6 & 9.593 & 6.089 & $\mathbf{4.225}$ & 5.546 \\
		\hline
		WF 7 & 0.867 & 0.838 & $\mathbf{0.677}$ & 0.612 \\
		\hline
		WF 8 & 8.042 & 5.656 & $\mathbf{3.451}$ & 3.865 \\
		\hline
		WF 9 & 4.713 & 5.511 & $\mathbf{3.060}$ & 3.405 \\
		\hline
		WF 10 & 1.875 & 3.117 & $\mathbf{1.025}$ & 1.424 \\		
		\hline
		Mean & 5.138 & 5.442 & $\mathbf{2.829}$ & 3.190 \\
		\hline
	\end{tabular}
	\label{tab:fitting_clear}
\end{table}
First, we analyze the fitting error results of the three CLeaR instances and the baseline model for the 11000 samples in the first two phases, see Tables~\ref{tab:fitting_ae} and~\ref{tab:fitting_clear}.
Table~\ref{tab:fitting_ae} shows the fitting errors regarding the weather-to-weather data, i.e., the outputs of the autoencoders, and Table~\ref{tab:fitting_clear} presents the corresponding results of the weather-to-power data, i.e., the outputs of the predictors.
WF is the abbreviation of the wind farm.
The values on the last row of the tables are the average results of the 10 wind farms.
The best results among the three CLeaR instances are marked in bold.
Compared to the baseline model results, Instance$_{\textbf{C}}$ can decrease the fitting error in the term of either input reconstruction or power prediction.
It is similar to our observation in experiment 1.
We can also observe that the average fitting error of the Instance$_{\textbf{B}}$ is sightly lower than that of the Instance$_{\textbf{C}}$ in Table~\ref{tab:fitting_ae}, but rises obviously in Table~\ref{tab:fitting_clear}.
We infer that Catastrophic Forgetting happens on the Instance$_{\textbf{B}}$ because it applies fine-tuning that can only adapt the model to the new situations.
We can obtain the conclusion regarding the existence of the forgetting problem according to the high forgetting ratio in Table~\ref{tab:forgetting} as well.

Table~\ref{tab:prediction_p} shows the results of power prediction, where the Instance$_{\textbf{C}}$ clearly outperforms the other two instances.
The baseline model results indicate that sufficient training data can enable neural network models to obtain as much meaningful knowledge as possible, which helps to improve the models' predictive ability.

\begin{figure}[t!]
	\includegraphics[width=0.49\textwidth]{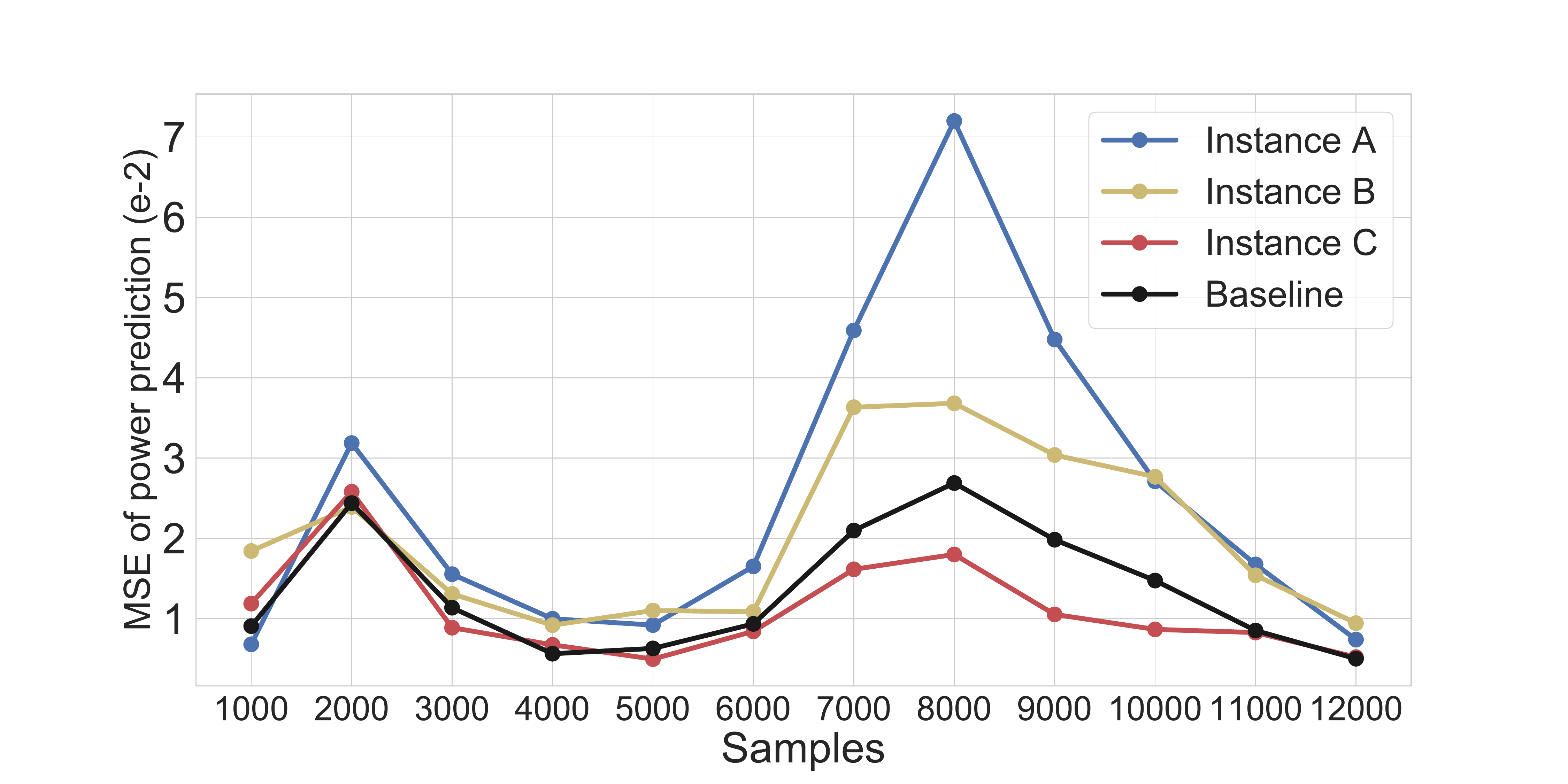}
	\caption{Errors of the three instances about one wind farm. Each point refers to the mean error over 1000 samples.}
	\label{fig:errors}
\end{figure}

Fig.~\ref{fig:errors} illustrates the errors of the three instances and the baseline model over the 12000 weather-to-power samples of one wind farm.
The errors are calculated after finishing the update phase.
We split the samples into 12 sections, and each point in Fig.~\ref{fig:errors} refers to an MSE of 1000 samples.
The average value of the points from 1000 to 11000 can be equal to the fitting error, and the value at a coordinate of 12000 is the prediction error.
Compared to the other three lines, the red line is less volatile.
Moreover, we can observe that the lines start to rise at a coordinate of 5000 and drop back to the original level at the end.
On the one hand, this reflects that the periodic changes can lead to these fluctuated curve shapes and influence the mapping between weather and power generation. 
On the other hand, in Tables~\ref{tab:fitting_clear},~\ref{tab:prediction_p}, and Fig~\ref{fig:errors}, where the Instance$_{\textbf{C}}$ outperforms the other instances, the results prove that updating is significant for predicting a non-stationary data stream, especially when the model is trained only on a limited dataset.
\begin{table}[t!]
	\caption{The prediction error of the \textbf{predictor} in the three CLeaR instances and the baseline model.}
	\begin{tabular}{c@{\hskip 3.5mm}c@{\hskip 3.5mm}c@{\hskip 3.5mm}c@{\hskip 3.5mm}|c@{\hskip 3.5mm}}
		\hline
		Dataset&\multicolumn{3}{c}{Prediction error (e-2)}\\ 
		\cline{2-5}
		&Instance$_{\textbf{A}}$ & Instance$_{\textbf{B}}$ & Instance$_{\textbf{C}}$ & Baseline\\
		\hline
		WF 1 & 2.415 & 2.332 & $\mathbf{1.822}$ & 1.611\\
		\hline
		WF 2 & 0.740 & 0.945 & $\mathbf{0.522}$ & 0.502\\
		\hline
		WF 3 & 2.370 & 5.369 & $\mathbf{1.208}$ & 1.641\\
		\hline
		WF 4 & 2.124 & $\mathbf{1.643}$ & 2.223 & 1.448\\
		\hline
		WF 5 & 13.864& 5.672 & $\mathbf{5.560}$ & 3.828\\
		\hline
		WF 6 & 8.612 & 8.017 & $\mathbf{4.368}$ & 3.123\\
		\hline
		WF 7 & 0.965 & 0.729 & $\mathbf{0.587}$ & 0.527\\
		\hline
		WF 8 & 6.129 & 4.969 & $\mathbf{2.927}$ & 2.499\\
		\hline
		WF 9 & 2.544 & 4.562 & $\mathbf{1.808}$ & 1.129\\
		\hline
		WF 10 & 1.383 & 3.009 & $\mathbf{0.741}$ & 0.866\\		
		\hline
		Mean & 4.115 & 3.725 & $\mathbf{2.177}$ & 1.717\\
		\hline
	\end{tabular}
	\label{tab:prediction_p}
\end{table}

Table~\ref{tab:forgetting} presents the forgetting ratio values between Instance$_{\textbf{B}}$ and Instance$_{\textbf{C}}$ calculated after the update phase. 
According to the average results, we can conclude that Instance$_{\textbf{C}}$ outperforms Instance$_{\textbf{B}}$ here, for both the weather-to-weather task and the weather-to-power task.
Moreover, as the conclusion of Table~\ref{tab:fitting_clear}, the existence of the forgetting problem leads to the increment of the fitting error.  
\begin{table}[t!]
	\caption{The forgetting ratio of Instance$_{\textbf{B}}$ and Instance$_{\textbf{C}}$.}
	\begin{tabular}{>{\centering}p{0.95cm}>{\centering\arraybackslash}p{1.3cm}>{\centering\arraybackslash}p{1.3cm}>{\centering\arraybackslash}p{1.1cm}>{\centering\arraybackslash}p{1.1cm}}
		\hline
		Dataset&\multicolumn{4}{c}{Forgetting ratio}\\ 
		\cline{2-5}
		&Instance$_{\textbf{B}_{ae}}$&Instance$_{\textbf{C}_{ae}}$ & Instance$_{\textbf{B}}$ &Instance$_{\textbf{C}}$\\
		\hline
		WF 1 & 1.474 & $\mathbf{0.727}$ & 1.955 & $\mathbf{0.994}$\\
		\hline
		WF 2 & $\mathbf{0.018}$ & 0.140 & 1.707 & $\mathbf{0.745}$\\
		\hline
		WF 3 & $\mathbf{1.205}$ & 1.489 & 6.170 & $\mathbf{0.421}$\\
		\hline
		WF 4 & 2.817 & $\mathbf{2.407}$ & 3.371 & $\mathbf{0.785}$\\
		\hline
		WF 5 & $\mathbf{0.000}$ & $\mathbf{0.000}$ & 7.617 & $\mathbf{2.296}$\\
		\hline
		WF 6 & 0.945 & $\mathbf{0.237}$ & 0.969 & $\mathbf{0.046}$\\
		\hline
		WF 7 & $\mathbf{0.574}$ & 1.552 & 1.768 & $\mathbf{1.253}$\\
		\hline
		WF 8 & $\mathbf{0.588}$ & 1.585 & 5.716 & $\mathbf{2.999}$\\
		\hline
		WF 9 & 1.463 & $\mathbf{0.848}$ & 3.470 & $\mathbf{1.463}$\\
		\hline
		WF 10 & 4.937 & $\mathbf{2.726}$ & 2.754 & $\mathbf{0.609}$\\		
		\hline
		Mean & 1.402 & $\mathbf{1.171}$ & 3.550 & $\mathbf{1.161}$\\
		\hline
	\end{tabular}
	\label{tab:forgetting}
\end{table}

\section*{Conclusions}
We believe that continual learning will be the key to the future machine's intelligence.
The non-stationary world requires future artificial intelligence to be updated smoothly by taking account into the different data distributions but still to retain previous useful knowledge.
Therefore, in this article, the proposed CLeaR describes the prototype structure of the continual learning based framework.
It can be applied to lots of real-world projects, such as power predictions for smart grids, where prediction models have to mimic humans' ability to acquire and transfer knowledge incrementally from new data throughout their lifespan.

The framework still needs to be improved in our future research. 
For example, although the cleaned exception is not considered, it is necessary to define and process such exceptions in a real application. 
Besides, we only use MSE to calculate the difference between predictions and measurements. 
However, this method might not work when the values are unavailable in unsupervised settings.
Therefore, we suggest estimating the uncertainty of new predictions to detect changes in distributions, for example, using Monte Carlo dropout~\cite{gal2016dropout}.
If the uncertainty is over a threshold, the data can be labeled as novelties. 
Futhermore, hyperparameters are found by grid search in the context of the current design. 
It is worth researching whether hyperparameters can be found by transferring relevant knowledge from a similar task dataset.
In addition, we suppose that Online-EWC can be replaced by (or combined with) other CL algorithms to improve the framework.
Novel analysis methods and evaluation metrics for the updated models will also be one of our main research focuses in future.

In a nutshell, we expect that the framework can be designed as a modular tool like LEGO toys. 
Each component of the framework is flexible and can be added, removed, replaced, or expanded. 
It should be possible for researchers and users to adapt the framework to a specific application scenario for achieving their own goals.


\begin{backmatter}
	
	\section*{Acknowledgements}
	Thanks to our colleagues from the Intelligent Embedded Systems group, particularly Mohammad Wazed Ali, Florian Heidecker, and Chandana Priya Nivarthi, for their helpful comments and suggestions. 
	
	\section*{Funding}
	This work was supported within the C/sells RegioFlexMarkt Nordhessen (03SIN119) project and the Digital-Twin-Solar (03EI6024E) project, funded by BMWi: Deutsches Bundesministerium für Wirtschaft und Energie/German Federal Ministry for Economic Affairs and Energy.
	
	\section*{Abbreviations}
	CL: Continual Learning;
	MSE: Mean Square Error;
	EWC: Elastic Weight Consolidation;
	SI: Synaptic Intelligence;
	LWF: Learning Without Forgetting;
	Online-EWC: Online Elastic Weight Consolidation;
	PCA: Principal Component Analysis;
	AE: Autoencoder;
	WF: Wind farm;
	
	\section*{Availability of data and materials}
	The datasets generated and analyzed during the current study are available from \url{https://www.uni-kassel.de/eecs/ies/downloads} or the corresponding author on reasonable request.
	
	\section*{Ethics approval and consent to participate}
	Not applicable
	
	\section*{Competing interests}
	The authors declare that they have no competing interests.
	
	\section*{Consent for publication}
	Not applicable
	
	\section*{Authors' contributions}
	YH wrote the majority of the manuscript and was responsible for implementing the CLeaR and experiments. BS suggested the artificial data and the first experiment and was responsible for double-checking the manuscript. All authors read and approved the final manuscript.
	
	\section*{Authors' information}
	Not applicable
	
	\bibliographystyle{bmc-mathphys} 
	\bibliography{bmc_article}      
	
	
	
	
\end{backmatter}
\end{document}